\documentclass[lettersize,journal]{IEEEtran}
\usepackage{amssymb}
\usepackage{graphicx,dsfont,amsmath,amsfonts}
\usepackage{subfigure}
\usepackage[section]{placeins}
\usepackage{stfloats}
\usepackage{float}
\usepackage{graphicx}
\usepackage{mathrsfs}
\usepackage{booktabs} % 导入booktabs包
\usepackage{multirow} % 导入multirow包

\usepackage[mathscr]{euscript}
\usepackage{xcolor}  % Include this for color support
\usepackage{hyperref}

\allowdisplaybreaks[4]
\makeatletter
\newcommand{\biggg}{\bBigg@{3}}

\makeatother
\makeatletter
\renewcommand{\@thesubfigure}{\hskip\subfiglabelskip}
\makeatother
\usepackage{cite}
\begin{document}

\title{SSFold: Learning to Fold Arbitrary Crumpled Cloth Using Graph Dynamics from Human Demonstration}

\author{Changshi Zhou, Haichuan Xu, Jiarui Hu, Feng Luan, Zhipeng Wang, \IEEEmembership{Member, IEEE}, Yanchao Dong, Yanmin Zhou, ~\IEEEmembership{Member, IEEE}, Bin He, \IEEEmembership{Senior Member, IEEE}

        % <-this % stops a space
\thanks{This work was supported in part by the National Natural Science Foundation
of China (No. 62088101, 62473294, 62403363), in part by the Science and
Technology Commission of Shanghai Municipality (No. 2021SHZDZX0100,
22ZR1467100), and the Fundamental Research Funds for the Central Universities
(No. 22120240291). (Corresponding author: Yanmin Zhou).}% <-this % stops a space
\thanks{Changshi Zhou, Haichuan Xu and Feng Luan are with the Shanghai Research Institute for Intelligent Autonomous Systems, and also with the National Key Laboratory of Autonomous Intelligent Unmanned Systems (Tongji University), Shanghai 201210, China, and also with the Frontiers Science Center for Intelligent Autonomous Systems, Shanghai 201210, China. (e-mail: 2210975@tongji.edu.cn; xuhaichuan@tongji.edu.cn; 2210957@tongji.edu.cn).}
\thanks{Jiarui Hu, Zhipeng Wang, Yanchao Dong, Yanmin Zhou and Bin He are with the College of Electronics and Information Engineering, Tongji University, Shanghai 201804, China, and also with the National Key Laboratory of Autonomous Intelligent Unmanned Systems (Tongji University), Shanghai 201210, China, and also with the Frontiers Science Center for Intelligent Autonomous Systems, Shanghai 201210, China. (e-mail: 2132976@tongji.edu.cn; wangzhipeng@tongji.edu.cn; dongyanchao@tongji.edu.cn; yanmin.zhou@tongji.edu.cn; hebin@tongji.edu.cn).}
\thanks{Color versions of one or more of the figures in this article are available online at http://ieeexplore.ieee.org}}

% \author{Anonymous Authors%
% \thanks{This paragraph of the first footnote will contain the date on which you submitted your paper for review, which is populated by IEEE. It is IEEE style to display support information, including sponsor and financial support acknowledgment, here and not in an acknowledgment section at the end of the article. 

% For example, “This work was supported in part by the National Natural Science Foundation of China (No. 62088101, 62473294, 62403363), in part by the Science and Technology Commission of Shanghai Municipality (No. 2021SHZDZX0100, 22ZR1467100), and the Fundamental Research Funds for the Central Universities (No. 22120240291).” 

% The name of the corresponding author appears after the financial information, e.g., (Corresponding author: Anonymous Author). Here you may also indicate if authors contributed equally or if there are co-first authors.

% The next few paragraphs should contain the authors’ current affiliations, including current address and e-mail. For example, Anonymous Author is with an Anonymous Institution (e-mail: anonymous@anonymous.com). 

% Mentions of supplemental materials and animal/human rights statements can be included here. 

% Color versions of one or more of the figures in this article are available online at http://ieeexplore.ieee.org.}%
% }
% \author{Anonymous Authors%
% }

% The paper headers
%\markboth{Journal of \LaTeX\ Class Files,~Vol.~14, No.~8, August~2024}%
%{Shell \MakeLowercase{\textit{et al.}}: A Sample Article Using IEEEtran.cls for IEEE Journals}

\IEEEpubid{}
% Remember, if you use this you must call \IEEEpubidadjcol in the second
% column for its text to clear the IEEEpubid mark.

\maketitle

\begin{abstract}
Robotic cloth manipulation poses significant challenges due to the fabric's complex dynamics and the high dimensionality of configuration spaces. Previous approaches have focused on isolated smoothing or folding tasks and relied heavily on simulations, often struggling to bridge the sim-to-real gap. This gap arises as simulated cloth dynamics fail to capture real-world properties such as elasticity, friction, and occlusions, causing accuracy loss and limited generalization. To tackle these challenges, we propose a two-stream architecture with sequential and spatial pathways, unifying smoothing and folding tasks into a single adaptable policy model. The sequential stream determines pick-and-place positions, while the spatial stream, using a connectivity dynamics model, constructs a visibility graph from partial point cloud data, enabling the model to infer the cloth's full configuration despite occlusions. To address the sim-to-real gap, we integrate real-world human demonstration data via a hand-tracking detection algorithm, enhancing real-world performance across diverse cloth configurations. Our method, validated on a UR5 robot across six distinct cloth folding tasks, consistently achieves desired folded states from arbitrary crumpled initial configurations, with success rates of 100.0\%, 100.0\%, 83.3\%, 66.7\%, 83.3\%, and 66.7\%.  It outperforms state-of-the-art cloth manipulation techniques and generalizes to unseen fabrics with diverse colors, shapes, and stiffness. Project page: \href{https://zcswdt.github.io/SSFold/}{https://zcswdt.github.io/SSFold/}
\end{abstract}

\def\abstractname{Note to Practitioners} 
\begin{abstract}
  In this paper, we introduce SSFold, a novel framework for robotic cloth manipulation that integrates human demonstrations with advanced learning techniques, providing a practical solution for real-world applications. Practitioners in industries such as textile manufacturing, automated laundry services, and even medical fabric handling can leverage this method to improve operational efficiency and reduce reliance on manual labor significantly. By using a two-stream architecture to handle complex cloth dynamics and self-occlusions, SSFold unifies smoothing and folding tasks into a single policy model, which adapts effectively to diverse fabric types and conditions. A key advantage of this method lies in its ability to utilize low-cost, easy-to-set-up hand-tracking systems for human demonstration data collection, reducing the need for expensive, complex setups. The framework has been successfully validated in real-world scenarios with a UR5 robot, achieving high success rates across a variety of folding tasks. Furthermore, the method demonstrates strong generalization capabilities, making it suitable for a wide range of applications beyond the tasks it was initially trained on. This scalability and flexibility offer a highly practical and cost-effective solution for integrating robotic cloth manipulation into various industries.

\end{abstract}


\section{Introduction}
\IEEEPARstart{C}{loth} manipulation has a wide range of applications in both domestic and industrial settings, such as laundry unfolding\cite{1} and folding\cite{2}, surgery\cite{3}, and manufacturing\cite{4}. These applications enhance the quality of life by reducing human labor. However, it has posed a challenge for robotic manipulation: compared to rigid objects, cloth has infinite degrees of freedom, can be only partially observable due to self-occlusions in crumpled configurations, and does not transform rigidly when manipulated. The dynamics of cloth are also complex\cite{5}, and slightly different interactions may lead to significantly different cloth behaviors. Early approaches for cloth manipulation efforts relied heavily on scripted actions, which were generally slow and lacked the flexibility to adapt to varying cloth configurations.

Recently, there have been two predominant learning-based approaches in the field of cloth manipulation. The first approach\cite{6}\cite{7}\cite{8}utilizes simulated data for training, specifically designed to capture the complex dynamics and varied states of cloth. While this method avoids the high costs of physical data collection, it struggles with significant sim-to-real gaps because many cloth states and dynamics are challenging to replicate accurately in simulators\cite{9}. The second approach\cite{10}\cite{11}directly collects human demonstration data from real-world interactions with cloth, facing challenges such as the high costs of data collection equipment. Therefore, to overcome the significant sim-to-real\cite{12} gaps and reduce the dependency on costly experimental setups for data collection in the real world, a new approach is essential that leverages more accessible and scalable methods of acquiring human demonstration data for robotic cloth manipulation.

\begin{figure*}[t]
  \centering
\includegraphics[width=0.95\linewidth]{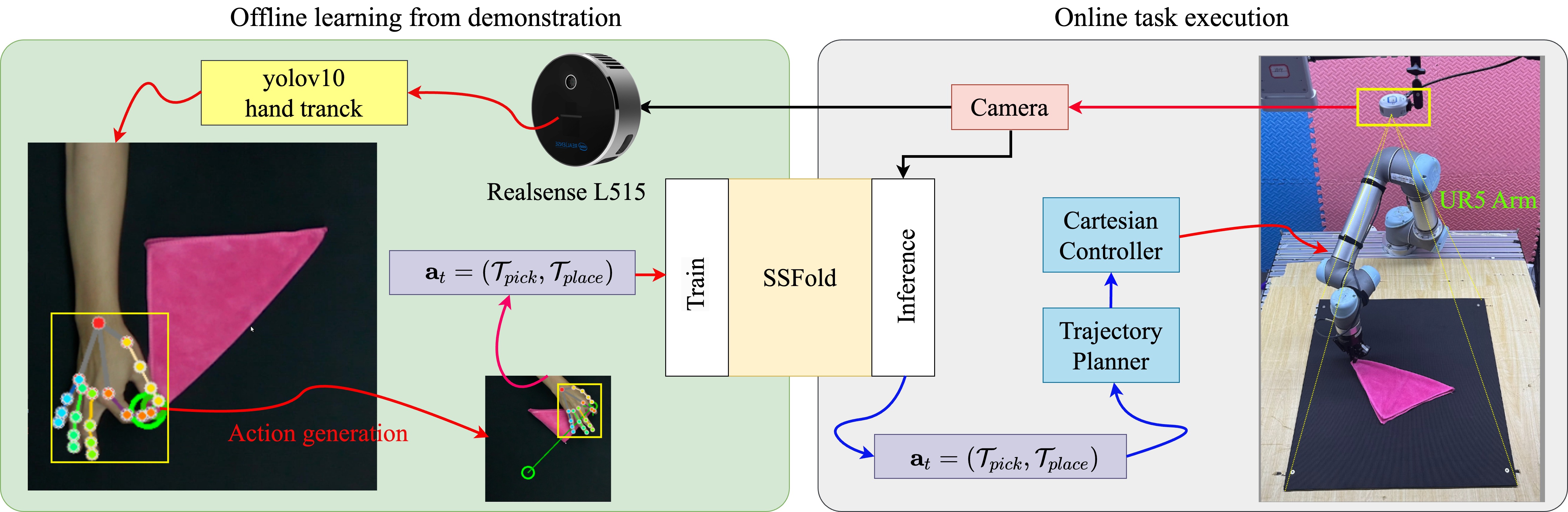}
\caption{\textbf{Fold Cloth from Human Demonstrations.} The proposed method comprises two stages: offline learning from demonstrations and online task execution. In the offline stage, human demonstration data is captured using hand-tracking techniques, and a neural network is trained and optimized iteratively. During the online stage, the neural network predicts poses from input images, which are then executed by the robot to perform similar actions.}
\label{}
\end{figure*}

In this paper, we introduce SSFold, an end-to-end policy model that utilizes human demonstration data to fold cloth from any initial crumpled configuration into the desired shape (see Fig.1). Our method leverages a YOLOv10-based\cite{13} system for hand tracking and key-point detection with a low-cost monocular camera, achieving accurate 3D estimation without the need for complex setups involving markers, hand-held devices, or multi-camera motion capture systems. To address the challenges of deformable dynamics, we establish a visible connectivity graph that effectively captures the spatial structure of the cloth, overcoming issues of partial observability and self-occlusions. Additionally, inspired by the pick-conditioned placing strategies of  Wu et al.\cite{14} and Thomas et al.\cite{15}, we incorporate this approach into our unified framework, extending its application beyond the original scope to effectively address both smoothing and folding tasks. Ultimately, the captured cloth states are input into SSFold, which then generates a distribution of action possibilities for picking and placing, pinpointing the optimal positions for these tasks based on the dynamics and configuration of the cloth. We validate our approach against three baselines\cite{10}\cite{11}\cite{15}, demonstrating significant performance improvements. Furthermore, extensive ablation experiments show that our method not only excels in real-world settings but also generalizes effortlessly to various cloth shapes and colors without additional training.

\noindent The contributions of this work are summarized as follows: 

We propose a YOLOv10-based method for hand tracking and key point detection to collect human demonstration data. Over 21 hours, we collected a dataset of 800 human demonstration videos, effectively bridging human-to-robot transfer by capturing intricate human manipulations for robotic training.

We propose a novel end-to-end two-stream architecture with sequential and spatial pathways, unifying smoothing and folding tasks into a single adaptable policy model.

Our method has been thoroughly evaluated using real robotic arms on a diverse range of cloth. Real-world experiments demonstrate that our approach outperforms state-of-the-art methods, with results showing it performs effectively in standardized settings and generalizes robustly to tasks unseen during the training phase.

The remainder of this paper is structured as follows. In Section II, we review the related work that forms the foundation of our study. Section III delves into the design of the proposed learning framework, providing a comprehensive explanation of its architecture and components. The experimental setup, along with a detailed analysis of the results, is presented in Section IV. Finally, Section V concludes the paper with a summary of our findings and discusses potential directions for future research.

\section{Related work}
\subsection{Human Demonstration}
Recent developments in Learning from Demonstrations (LfD) have increasingly embraced learning-based methods, particularly in the realm of cloth manipulation, where human demonstrations have been shown to significantly improve performance\cite{16}\cite{17}. Traditionally, LfD relied on teleoperation techniques using devices such as joysticks or VR interfaces\cite{18}\cite{19} and kinesthetic teaching, where operators manually guide robot arms\cite{20}\cite{21}. While these methods have proven effective, they typically involve complex and costly setups, including motion capture systems and sensor-equipped gloves, which can restrict natural human movements and complicate the data collection process, thus limiting their scalability and practicality in diverse settings. To overcome these challenges, there has been a shift towards utilizing advancements in human pose estimation to refine robotic learning from human demonstrations. This newer approach leverages videos that capture direct human-environment interactions, allowing researchers\cite{22}\cite{23}\cite{24} to teach robots by extracting key actions and dynamics from these demonstrations. By eliminating the need for elaborate and expensive setups, human pose estimation not only simplifies the data collection process but also reduces costs. More importantly, it captures the essential dynamics with high accuracy, significantly enhancing the robot’s ability to learn effectively from human examples. This evolution in data collection and training methods has rendered robotic training more practical, cost-effective, and broadly applicable to various real-world scenarios. Building on this progress, our approach employs a YOLOv10-based system and a monocular camera to efficiently collect high-quality human demonstration data. This method avoids the complexities of traditional systems and enhances the flexibility and applicability of data collection for robotic training, making it more adaptable to real-world settings.

\begin{figure*}[t]
\centering
\includegraphics[width=1\linewidth]{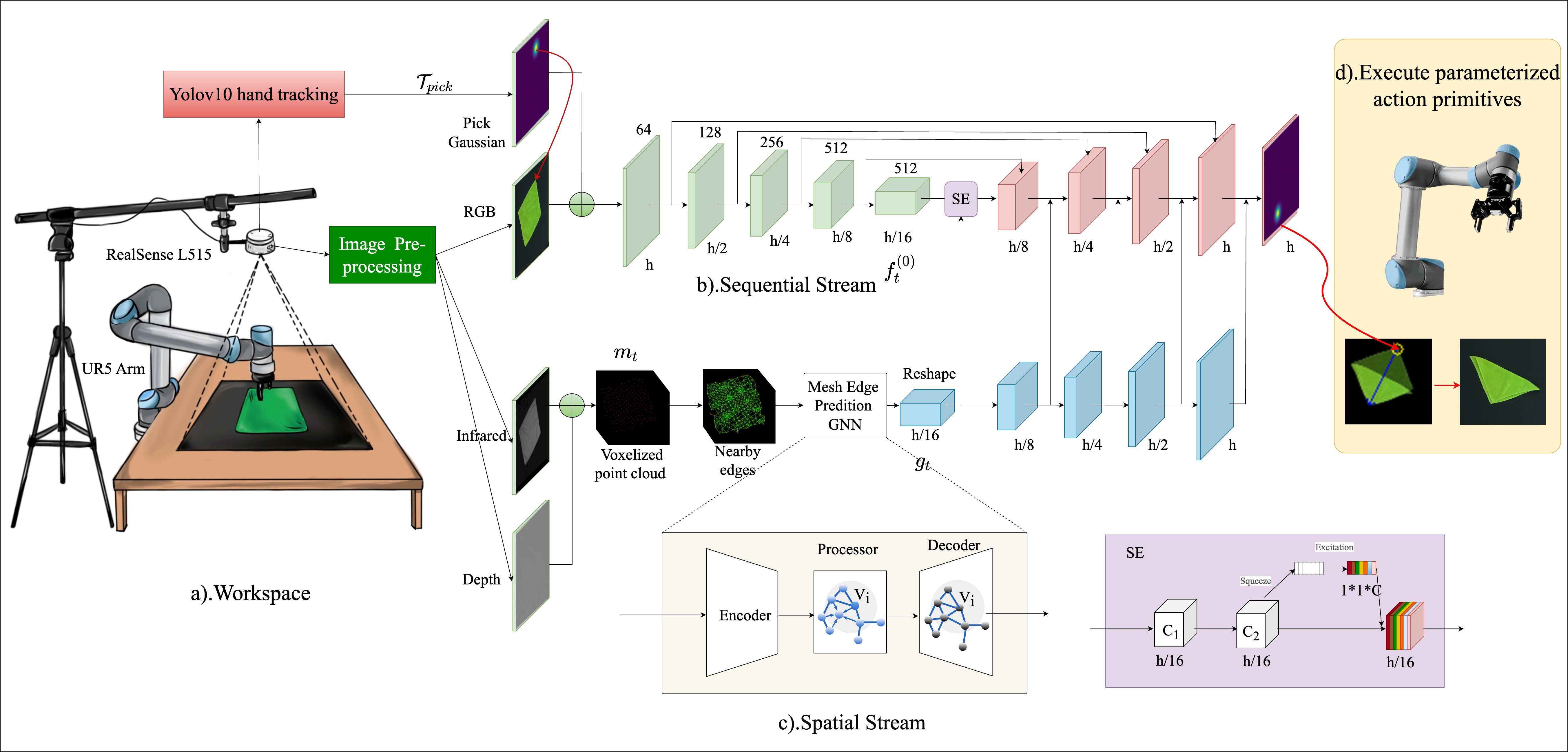}
\caption{\textbf{Method Overview.} (a) In a workspace equipped with a UR5 robotic arm and a piece of cloth in an arbitrary crumpled configuration, a top-down RGB image is captured by the camera. (b) The pick point, identified using a YOLOv10-based hand tracking algorithm, is concatenated with the captured RGB image. This combined input is then fed into the U-net network within the Sequential Stream. (c) In the Spatial Stream, the infrared and depth images captured by the camera are first used to extract a mask of the cloth region and generate the corresponding point cloud. The point cloud is voxelized to reduce complexity, followed by inferring nearby edges and mesh edges to predict the cloth's graph data. (d) Finally, the features from both streams are fused and processed to produce an output action map, which guides the robotic arm to execute the corresponding actions using parameterized action primitives.}
\end{figure*}

\subsection{Cloth Smoothing}
Cloth smoothing aims to transform the cloth from an arbitrary crumpled configuration to a smooth configuration\cite{25}. Prior work has focused on extracting and identifying specific features such as corners\cite{26} and wrinkles\cite{27}. Jianing et al.\cite{28} used a segmentation network to identify the edges of towels for flattening. However, this method loses effectiveness when the edges are occluded. Recent learning-based methods have used expert demonstrations\cite{2}\cite{29} to learn cloth smoothing policies in simulation\cite{30}\cite{25}(e.g., Pyflex\cite{9}) for data collection. David et al. \cite{31} employed Softgym\cite{32} as the simulation environment, utilizing the Quasi-Dynamic Parameterisable (QDP) method, which optimized parameters such as motion velocity for cloth smoothing. Zixuan et al. \cite{33} proposed a Mesh-based Dynamics approach capable of performing both cloth flattening and canonicalization. However, many complex cloth states, materials, and dynamics cannot be accurately modeled by these simulators. Thus, the sim-to-real gaps are significant obstacles for these methods to achieve better generalization in real-world applications. In comparison, we directly collect human demonstration data and learn from these real-world interactions to effectively model the diverse and intricate behaviors of cloth. By leveraging human intuition and dexterity, our approach can capture complex cloth dynamics, including variations in material properties and folding patterns that are often difficult to replicate in simulation environments. This data-driven learning significantly narrows the sim2real gaps and enhances the robustness and accuracy of our method in real-world applications, ensuring better generalization across a wide variety of cloth types and conditions.

\subsection{Cloth Fold}
Cloth folding has numerous applications in hospitals, homes, and warehouses. Early approaches heavily relied on heuristic-based methods, achieving high success rates but limited generalization due to strong assumptions about cloth types, textures, and shapes. Recent methods have focused on learning goal-conditioned policies\cite{15}\cite{34}\cite{35} in simulation. Lee et al\cite{10} developed a goal-conditioned pick-and-place policy using a Deep Q-network and self-supervised real-robot experience. However, the model's performance was limited by the simplicity of the network architecture and the insufficient quality of the robotic experience. Robert et al. [11] proposed a fully convolutional network conditioned on pick locations for cloth folding, trained on a small dataset for a single task. The model’s performance was hindered by a lack of solutions for self-occlusion and the limited diversity of the training data. Thomas et al. \cite{15} introduced an optical flow-based approach to represent actions for fabric folding, which improved performance in certain conditions. However, when the cloth configuration deviated from the demonstration sub-goals, optical flow estimation became inaccurate, leading to folding failures. Recent work, such as \cite{36}, learns dense visual correspondences for category-level garment manipulation but suffers from a sim-to-real gap. Additionally, \cite{37} enables garment folding using an assistive tool but is limited to folding tasks and requires extra hardware, restricting its flexibility. In contrast, our SSFold framework combines both smoothing and folding in a single end-to-end model. It can flatten cloth from any initial state and fold it into various target shapes, all while avoiding the sim-to-real gap.
\section{Methods}

\subsection{Problem Formulation}
This study aims to develop a framework for robotic cloth manipulation, specifically focusing on learning from expert demonstrations to flatten crumpled cloth and fold it into target shapes. At time $t$, we obtain a top-view visual observation of the current cloth $o_t \in \mathbb{R}^{W \times H \times C}$. The robot's actions, defined by picking and placing poses $\mathbf{a}_t=\left(\mathcal{T}_{\text {pick }}, \mathcal{T}_{\text {place }}\right)$, involve grasping the cloth at $\mathcal{T}_{\text {pick }}$, lifting it slightly, dragging it parallel to the workspace plane towards $\mathcal{T}_{\text {place }}$, and releasing it. Both $\mathcal{T}_{\text {pick }}$ and $\mathcal{T}_{\text {place }}$ are 3D world coordinate positions, derived from 2D image pixels via robot-camera calibration. 

We formulate the problem as learning a goal-conditioned policy $\pi$ that generates actions $a_t$ based on the input $\gamma_t=$ $\left(o_t, g_t\right)$. The visual observation $o_t$ is combined with a representation graph $g_t$, which encodes the spatial relationships and features of the cloth obtained from depth sensors and processed through a graph neural network. The policy is defined as shown in Equation (1):
\begin{align}
\pi\left(\boldsymbol{\gamma}_t\right)=\pi\left(\mathbf{o}_t, \mathbf{g}_t\right) \rightarrow \mathbf{a}_t=\left(\mathcal{T}_{\text {pick }}, \mathcal{T}_{\text {place }}\right) \in \mathcal{A}
\end{align}

\subsection{Learning Pick-Conditioned Place}

The evolution of manipulation strategies for cloth has progressed through distinct stages. Initially, methodologies employed a unified network to simultaneously predict both pick and place points. However, this approach did not optimally exploit the sequential nature of these actions, potentially leading to suboptimal coordination between picking and placing. Weng et al.\cite{15}refined the approach by implementing a dual-network strategy, featuring two distinct $Q$-functions. The pick module $Q_{\text {pick }}$ identifies the optimal pick location, while the place module $Q_{\text {place }}$, conditioned on the pick decision, determines the best placement action. This separation into specialized networks enhances precision and allows for the optimization of each action based on the previous one's results. Building upon prior developments, our methodology (see Fig.2(b)) advances by using a single network to condition place actions directly on pick actions. This approach simplifies the architecture while maintaining the essential relationship between picking and placing, enhancing precision in fabric manipulation.

During the training phase, the ground truth for each pick point is transformed into a two-dimensional Gaussian distribution and combined with $o_t$ features before being fed into the network to predict place positions $\mathcal{T}_{\text {place. }}$. In the testing phase, the model selects $N$ potential pick points from the segmented cloth mask, transforms each into a two-dimensional Gaussian distribution, merges them with $o_t$ features, and this data is then  fed into the network to produce $N$ $\mathit{place\_maps}$. The optimal $\mathit{place\_map}$ is identified by determining the index $i$ that maximizes the probability across all $\mathit{place\_maps}$, conditioned on their respective pick points:
\begin{align}
i^*=\arg \max _i\left(\max _{(u, v)} P_i\left((u, v) \mid \mathcal{T}_{\text {pick }_i}\right)\right)
\end{align}

Following this, the most probable place position $\mathcal{T}_{\text {place }}$ within the optimal placemap $P_{i^*}$ is identified through:
\begin{align}
\mathcal{T}_{\text {place }}=\arg \max _{(u, v)} P_{i^*}\left((u, v) \mid \mathcal{T}_{\text {pick }_{i^*}}\right)
\end{align}

Upon determining $\mathcal{T}_{\text {place }}$, the corresponding $\mathcal{T}_{\text {pick }}$ is traced, thereby completing the prediction of the pick.

\begin{figure}[t]
  \centering
\centering
\includegraphics[width=3.5in]{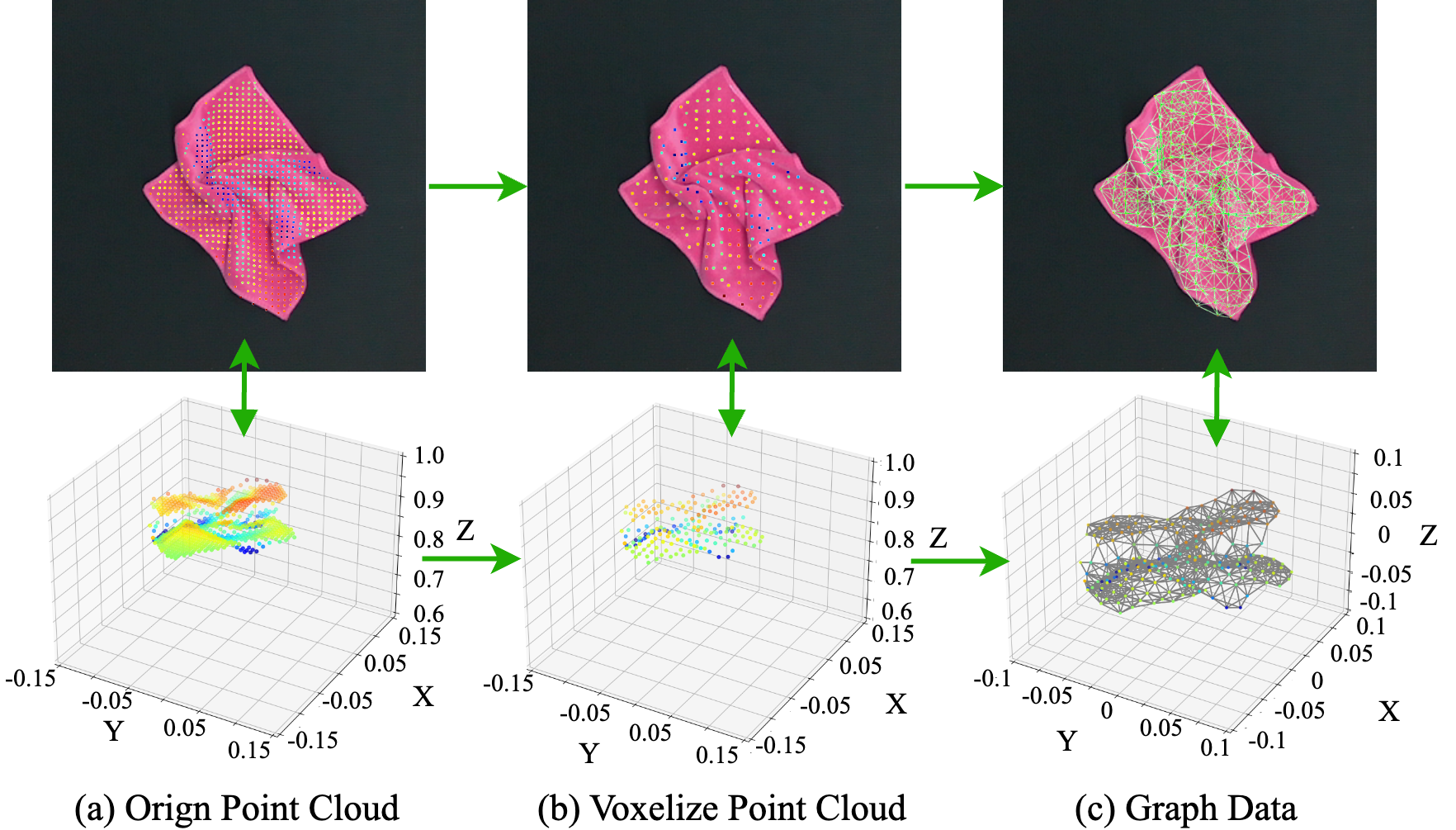}
%\caption{fig2}
\caption{\textbf{Graph Construction for Cloth Particles.} Initially, the point cloud data of the cloth(a) undergoes voxelization to generate (b). Then, applying methods from Section III(C), particle connections are predicted to create graph(c).}
\label{}
\end{figure}

\subsection{Learning Graph Dynamics}
To tackle the challenges of self-occlusion in cloth manipulation, we introduce a visible connectivity graph \( G_t = \langle V, E \rangle \) to infer connections among visible points on the cloth mesh. The nodes \( V \) in the graph correspond to particles derived from the voxelized point cloud, while the edges \( E \) represent both the intrinsic structural connectivity of the cloth and its dynamic interactions during manipulation (see Fig. 3).

\textbf{Point Cloud Voxelization. }Depth images of self-occluded cloth, captured by an RGB-D camera, are transformed into 3D coordinates using the camera’s intrinsic and extrinsic parameters. These coordinates form the raw point cloud \( P_{\text{raw}} = \{ x_i \}_{i=1}^{N_{\text{raw}}} \), where \( x_i \) represents the position of each point in 3D space, and \( N_{\text{raw}} \) is the total number of points. While the raw point cloud provides a detailed spatial representation of the cloth surface, it often exhibits uneven density and noise, making it unsuitable for direct graph construction.

To address these issues, \( P_{\text{raw}} \) undergoes a voxelization process. A uniform 3D voxel grid is overlaid onto the point cloud, dividing the space into cubic regions with a fixed edge length \( l \). For each voxel, the centroid of all points within the voxel is computed:

  \begin{equation}
    x_{\text{centroid}} = \frac{1}{n} \sum_{j=1}^{n} x_j
    \end{equation}

    \noindent where \( n \) is the number of points in the voxel, and \( x_j \) is the position of each point. This process produces a voxelized point cloud \( P = \{ x_{\text{centroid}, i} \}_{i=1}^{N_p} \), where \( N_p \) is the number of voxels. Each point \( x_{\text{centroid}, i} \) represents a single voxel and serves as a node in the subsequent graph construction step.

\textbf{Graph Construction. }After voxelizing the raw point cloud \( P_{\text{raw}} \), the resulting voxelized points are used to construct the visible connectivity graph \( G_t = \langle V, E \rangle \), where each voxelized point \( x_{\text{centroid}, i} \) corresponds to a node \( v_i \), representing a cloth particle. The graph edges \( E \) are classified into mesh edges \( E_M \) and nearby edges \( E_C \).

Mesh edges \( E_M \) represent the fixed connections between particles in the underlying cloth mesh. These edges are determined by the cloth’s structure and remain constant over time. Each edge \( e_{ij} = (v_i, v_j) \in E_M \) connects nodes \( v_i \) and \( v_j \), modeling the inherent mesh connectivity between them. Nearby edges \( E_C \), on the other hand, model the collision dynamics between particles during cloth deformation. As the cloth bends or folds, particles may come close to each other, even if not connected by a mesh edge. To capture these interactions, nearby edges are dynamically constructed at each time step based on the following proximity criterion:

\begin{equation}
    E_C = \left\{ e_{ij} \mid \| x_i - x_j \|_2 < R \right\}
    \end{equation}

    \noindent where \( \| x_i - x_j \|_2 \) is the Euclidean distance between nodes \( i \) and \( j \), and \( R \) is a distance threshold.

However, due to the partial visibility of the cloth and the presence of self-occlusions, mesh edges cannot be directly observed from the point cloud alone. Instead, these edges are inferred through a pre-trained Edge Graph Neural Network (GNN)\cite{7}. By combining both mesh edges and nearby edges, the visible connectivity graph effectively captures both the cloth’s underlying static structure and its dynamic interactions during manipulation.

\subsection{Goal-Conditioned Network Architecture }
We choose a U-Net architecture over the fully convolutional net used in prior robotics manipulation work\cite{1}, as U-Nets are better suited for high-resolution inputs necessary for detecting edges and wrinkles for cloth smoothing. We extend the U-Net to two pathways: sequential and spatial (See Fig. 2).

This sequential stream replicates the U-Net architecture and is specifically designed to process RGB inputs $o_t$. It outputs dense features through an hourglass encoder-decoder model up to the last layer, represented as $o_t \rightarrow f_t^{(0)}, f_t^{(0)} \in \mathbb{R}^{4 \times 4 \times 1024}$. In this stream, the features are decoded and unsampled across subsequent layers $f_t^{(l-1)} \rightarrow f_t^{(l)}, f_t^{(l)} \in \mathbb{R}^{h \times w \times C}$, and are conditioned with a two-dimensional Gaussian distribution centered on the pick coordinates. At the bottleneck, these features are integrated with intermediate features from the spatial stream.

The spatial stream extracts the cloth's mask from infrared (IR) and depth images. The mask is used to reconstruct the cloth's 3D point cloud utilizing both intrinsic and calibrated extrinsic camera parameters. This point cloud is then voxelized to reduce complexity and improve computational efficiency, providing a structured grid that simplifies further processing. Utilizing this structured voxelized data, nearby edges of the cloth are computed according to Equation 5. A pre-trained edge GNN then is employed to predict the mesh edges $E^M$ from the $\operatorname{graph}\left\langle\mathbf{P}, E^C\right\rangle$.

The edges are encoded into a high-dimensional feature space $m_t \rightarrow g_t, g_t \in \mathbb{R}^{1024}$, capturing detailed spatial interactions. These features $g_t$ are then reshaped and passed through decoding layers, which upscale the features to match the dimensions used in the sequential stream, enhancing the congruence between the spatial and sequential data. To integrate with the sequential data, $g_t$ is downsampled using fully-connected layers to match the channel dimension C and tiled to align with the spatial dimensions of the sequential decoder features, resulting in $g_t \rightarrow g_t^{(l)}, g_t^{(l)} \in \mathbb{R}^{h \times w \times c}$. We utilize the SE module to fuse features ${f}_t^{(0)}$ and $g_t$, which enhances the network's ability to dynamically prioritize the most informative aspects, thereby improving overall performance. Finally, we integrate lateral connections between the spatial and semantic streams. These connections involve concatenating two feature tensors and applying $1 \times 1$ conv to reduce the channel dimension: 
\begin{align}
\left[\mathbf{f}_t^{(l)}; \mathbf{g}_t^{(l)}\right]: \mathbb{R}^{h \times w \times c_{\mathbf{v}}+c_{\mathbf{d}}} \rightarrow \mathbb{R}^{h \times w \times c_{\mathbf{v}}}
\end{align}
where $\mathbf{f}_t^{(l)}$ and $\mathbf{g}_t^{(l)}$ are the sequential  and spatial tensors at layer $l$, respectively.

  \begin{figure}[t]
    \centering
  \centering
  \includegraphics[width=1\linewidth]{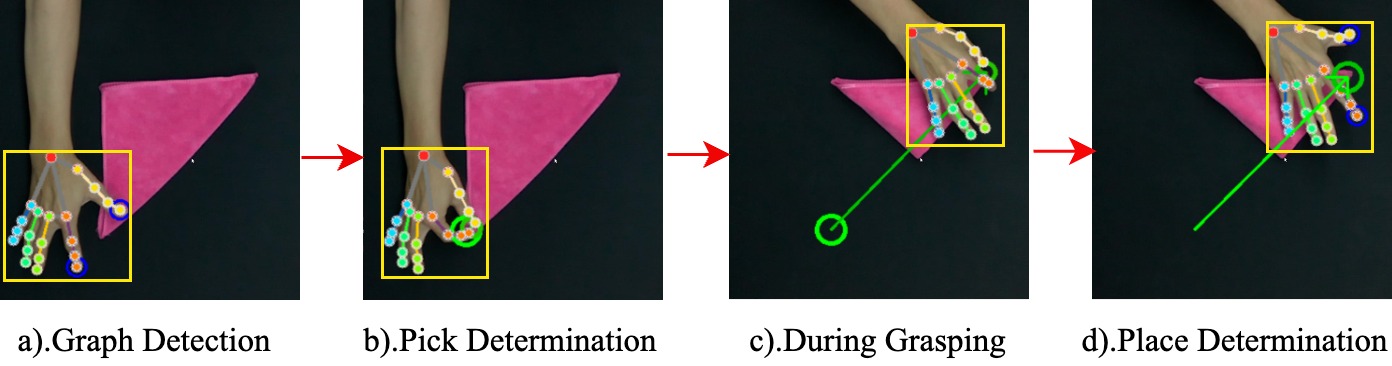}
  %\caption{fig2}
  \caption{\textbf{Human Demonstration Data Collection Process.} (a)Hand Pose Detection. (b) The pick point is determined when the thumb and forefingers meet. (c) Tracking the movement of the hand. (d) The place point is determined when the thumb and forefingers separate.}
  \label{}
  \end{figure}

\subsection{YOLOv10-Based Human Demonstrations Data Collection}

YOLOv10 is a state-of-the-art object detection framework that performs both bounding box prediction and class classification in a single forward pass, enabling real-time detection and tracking of objects. By introducing a dual assignment strategy, YOLOv10 eliminates the need for Non-Maximum Suppression (NMS), which not only speeds up the detection process but also enhances accuracy, particularly for small, fine-grained objects like the thumb and index finger in cloth manipulation tasks.

In this work, we adopt a pick-and-place action space for manipulation, focusing on recording the pick and place positions during human cloth manipulation. The thumb and index finger are used due to their dexterity and common use in grasping objects like cloth. YOLOv10 tracks the positions of these fingers in each frame of the RGB video streams recorded during human demonstrations. The framework generates bounding boxes around the hand and provides the coordinates of the keypoints for the thumb and index finger, allowing us to track their spatial positions across frames. This tracking data is then used to detect grasping and releasing actions.

Grasping is detected when the distance between the two fingers falls below a predefined threshold, while a release action is identified when this distance exceeds the threshold. The average position of the thumb and index finger is recorded as the pick or place location, respectively. These positions are stored as tuples of pixel coordinates, directly corresponding to the human's manipulation actions (see Fig. 4). To ensure accurate recording of the cloth’s state before and after manipulation, we analyze the video stream for periods of minimal pixel change, indicating that the hand has moved out of the frame and the cloth is stationary. During these intervals, static images are captured to document the cloth's state. Another image is captured once the manipulation is complete.

To enable precise hand tracking, we trained a YOLOv10-based hand-tracking model using 62,000 annotated public hand images. This trained model was then applied to collect demonstration data across various manipulation tasks. The use of infrared images helped isolate cloths from the foam background, facilitating more accurate tracking of fabrics with different colors. This led to the collection of infrared, depth, and RGB data, resulting in 800 demonstration episodes, including towels, T-shirts, and shorts, each starting from a crumpled initial state.

\begin{figure}[t]
  \centering
\centering
\includegraphics[width=1.6in]{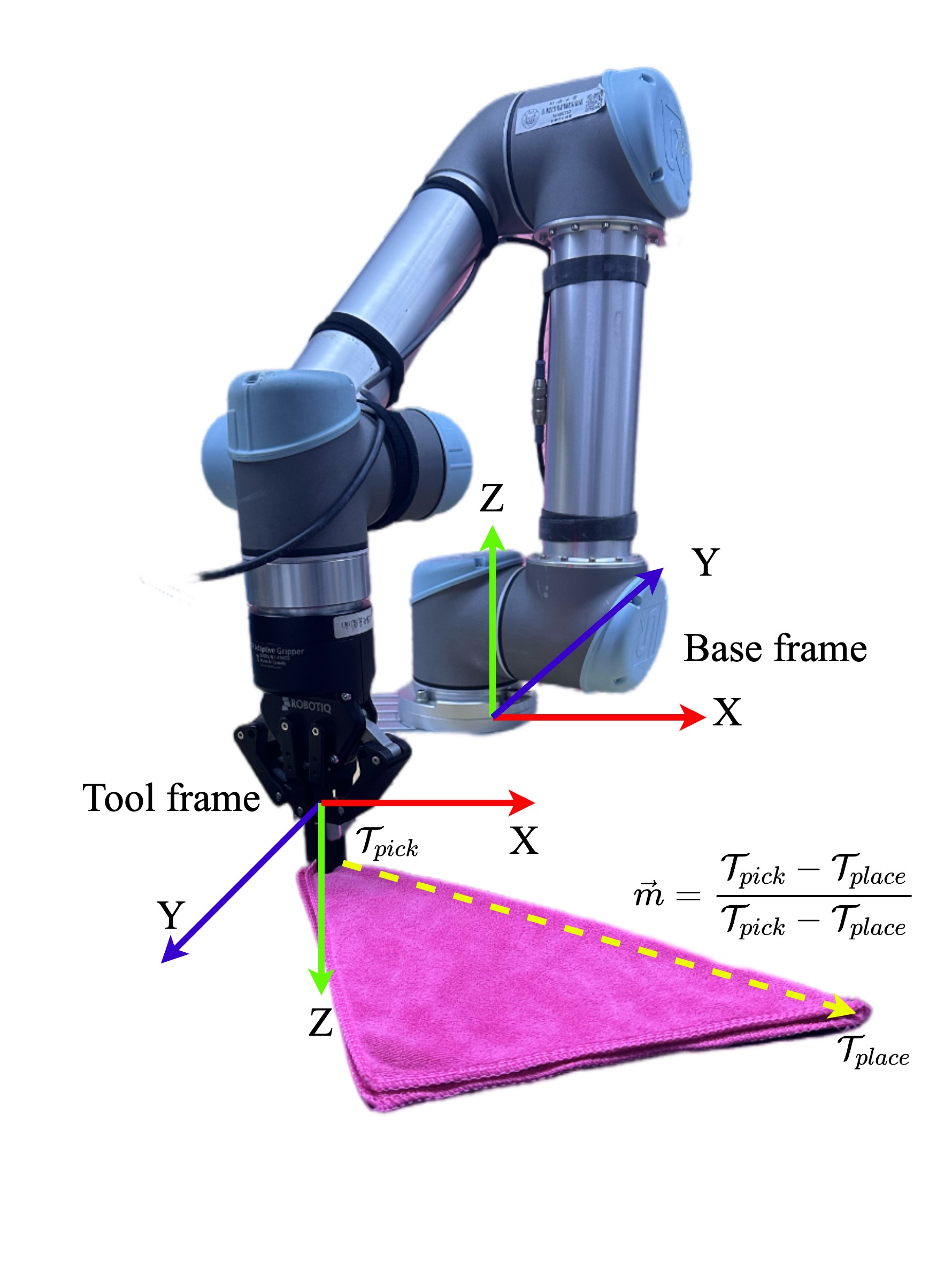}
%\caption{fig2}
\caption{\textbf{Gripper Direction Optimization. } The gripper's grasping direction is consistently aligned with the vector from the pick point to the place point.}
\label{}
\end{figure}

\subsection{Gripper Direction}
To optimize gripper orientation for robotic cloth manipulation, the directional vector from $\mathcal{T}_{\text {pick }}$ to $\mathcal{T}_{\text {place }}$ is normalized to unit length (See Fig. 5). This adjustment ensures that the gripper approaches the cloth according to its inherent lay direction, thus minimizing deformation:
\begin{align}
\mathbf{x}=\frac{\mathcal{T}_{\text {pick }}-\mathcal{T}_{\text {place }}}{\left\|\mathcal{T}_{\text {pick }}-\mathcal{T}_{\text {place }}\right\|}
\end{align}

A right-handed coordinate system, essential for consistent robotic operation, is established by setting the z -axis downward and computing an orthogonal y-axis: 
\begin{align}
\mathbf{y}&=\frac{[0,0,-1] \times \mathbf{x}}{\|[0,0,-1] \times \mathbf{x}\|} \\
\mathbf{x}&=\mathbf{y} \times[0,0,-1]
\end{align}

This configuration forms a rotation matrix $\mathbf{R}$ which is converted into a rotation vector $r$. This vector is key to the robotic control system, ensuring minimal deformation during gripping, thus improving cloth handling quality and durability.
\begin{align}
  \mathbf{R}=[\mathbf{x}~\mathbf{y}~[0,0,-1]]
  \end{align}

  \begin{figure}[t]
    \centering
    \includegraphics[width=1\linewidth]{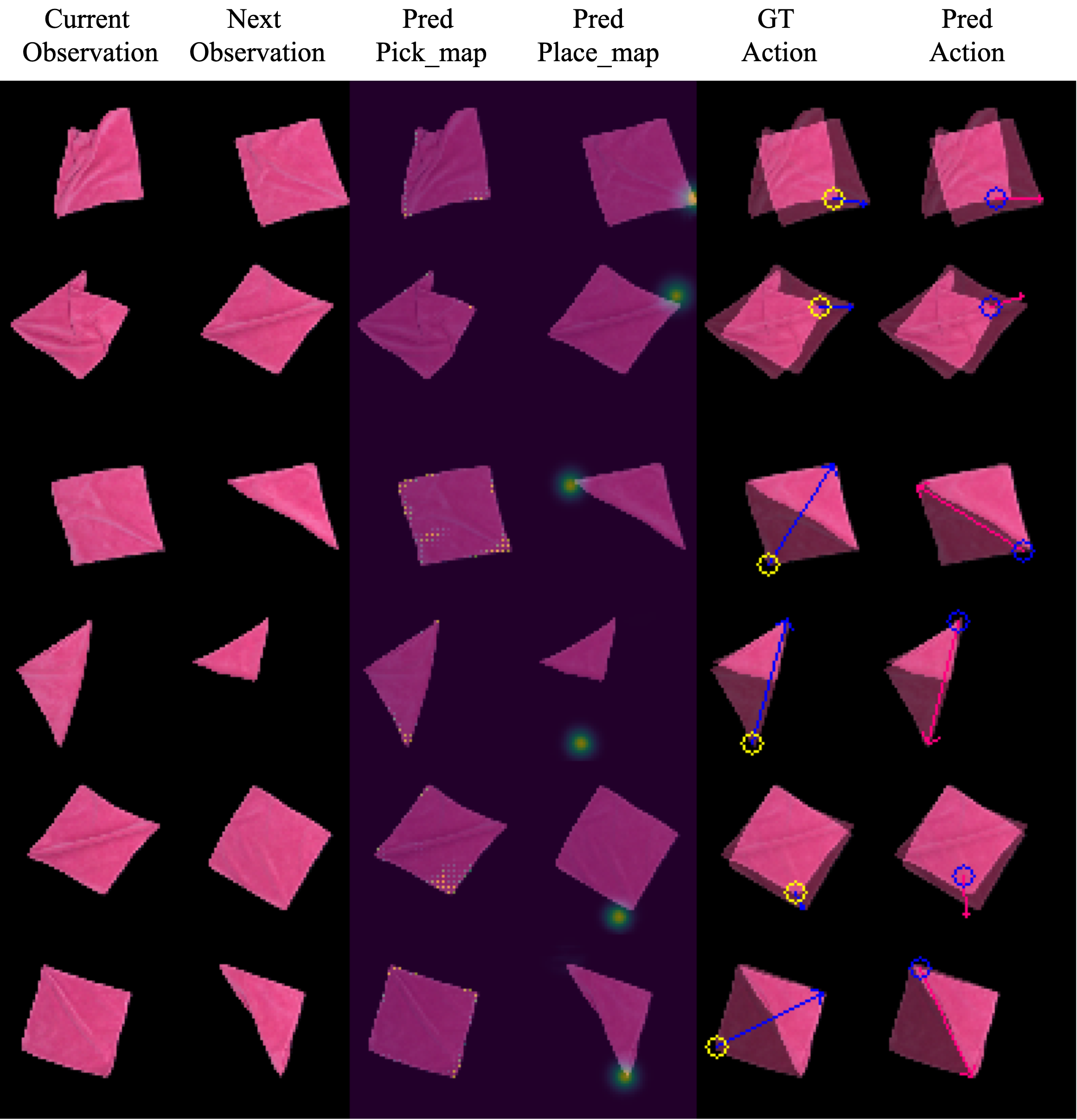}
    \caption{\textbf{Model Evaluation.} The trained model was evaluated on the test dataset using a double triangle fold task. The first column shows the current observation. The second column presents the next state resulting from executing the action $a_t$, which is determined based on the model's predicted \textit{pick\_map} (third column) and \textit{place\_map} (fourth column). The fifth column displays the ground truth action, and the sixth column shows the actual resulting action applied to the cloth. In the case of the square cloth and the double triangle fold task, the fold can be achieved through multiple valid action sequences. Therefore, if the predicted and actual actions result in the same final folded shape, it is considered a successful prediction, even if the specific action sequence differs.}
    \end{figure}

\subsection{Training Details}
In this study, we employ imitation learning to train our models using a dataset comprised of $N$ expert demonstrations, denoted as $\mathcal{D}=\left\{\zeta_1, \zeta_2, \ldots, \zeta_n\right\}$. Each trajectory $\zeta_i=$ $\left\{\left(\mathbf{o}_1, \mathbf{g}_1, \mathbf{a}_1\right),\left(\mathbf{o}_2, \mathbf{g}_2, \mathbf{a}_2\right), \ldots\right\}$ is a sequence of one or more observation-action pairs $\left(\mathbf{o}_t, \mathbf{g}_t, \mathbf{a}_t\right)$, where $\mathbf{o}_t$ represents the observation, $\mathbf{g}_{\mathrm{t}}$ the graph, and $\mathbf{a}_t$ the corresponding action at time t. For each task listed in Table II, we uniformly sample observation-action pairs from the dataset $\mathcal{D}$. Actions $\mathbf{a}_t$ are decomposed into two distinct training labels: $\mathcal{T}_{\text {pick }}$ and $\mathcal{T}_{\text {place }}$, which serve to generate binary one-hot pixel maps $Q_{\text {pick }} \in$ $\mathbb{R}^{H \times W}$ and $Q_{\text {place }} \in \mathbb{R}^{H \times W \times k}$, respectively. Initially, the edge GNN is frozen to ensure the stabilization of spatial relationship learning. Subsequently, we train the remaining modules by applying the Binary Cross-Entropy (BCE) loss between the predicted placement heatmaps $Q_{\text {place }}$ and the ground truth $Q_{\mathrm{place}}^{\mathrm{gt}}$ computed as:
\begin{align}
\mathcal{L}=\operatorname{BCE}\left(Q_{\text {place }}, Q_{\text {place }}^{\mathrm{gt}}\right)
\end{align}
The model is trained with a batch size of 32, the Adam optimizer, a learning rate of 0.0001, and 30,000 iterations. The process is executed on an NVIDIA RTX 4090 GPU. Fig. 6 displays the performance of the trained model on the test set.

\section{Experiments}
\subsection{Real World Setup}
Fig.2(a) illustrates the robotic system set up in a real-world environment. We utilize a 6DOF UR5 robotic arm outfitted with Robotiq 2-Finger Grippers, positioned on a 1 m x 1 m workspace table. Trajectories for pick and place operations are planned using URScript. A rigid Polypropylene (PP) board table is situated within the workspace to enable the grippers to reach beneath the cloth without collision. Directly above, an Intel RealSense L515 camera captures RGBD images, generating observations $o_t \in \mathbb{R}^{640 \times 480 \times 4}$, providing both color and depth images for point cloud construction. Computing is done on a system using an Intel i5-13900K CPU, 64GB RAM, and an NVIDIA GeForce RTX 4090.

\begin{figure}[t]
  \centering
\centering
\includegraphics[width=1\linewidth]{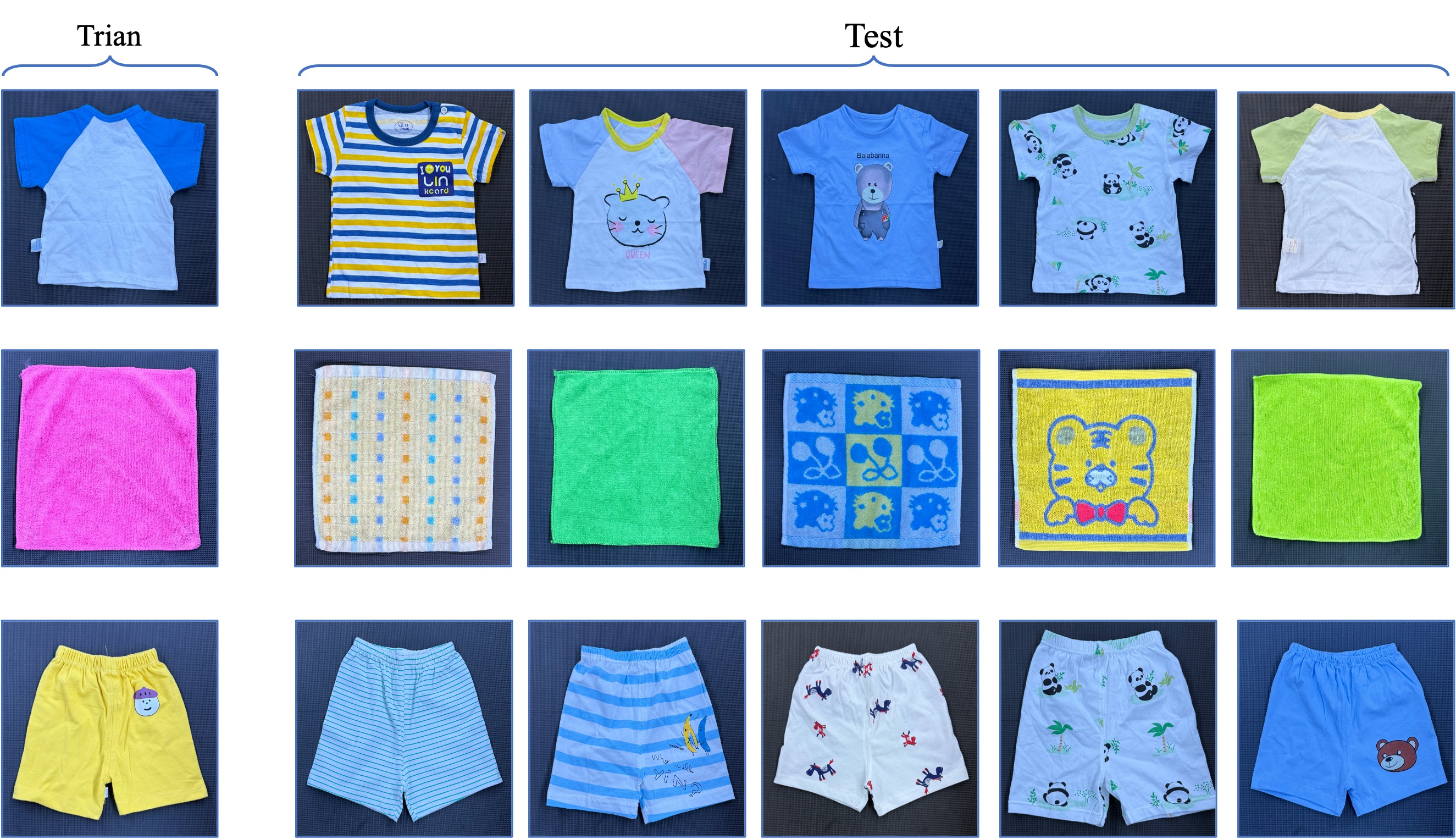}
%\caption{fig2}
\caption{\textbf{Fabric Categories Used for Model Training and Testing.} The dataset includes 18 samples from each of three fabric types: towels, T-shirts, and shorts.}
\label{}
\end{figure}

\subsection{Tasks and Metrics}

\textbf{Intermediate Smooth Coverage (ISC):}  This is the ratio of fabric coverage (sum of mask pixels) to the coverage of a smoothed cloth mask, with the highest coverage score across all timesteps in the episode used to assess the degree of fabric smoothing.

\textbf{Success(S):} This metric assesses the quality of smoothing and folding by determining whether the final state visually matches the goal image.

\textbf{Actions(A):}  This represents the number of actions taken during the manipulation process.

\textbf{Delta Coverage ($\Delta$):} This is Calculated as the final coverage minus the initial coverage, representing the change.

\textbf{Intersection over Union (IoU):} This metric measures the quality of cloth smoothing by comparing the manipulated cloth's mask with the target mask.

\textbf{Grasping Accuracy (GA):}The model’s ability to accurately predict the grasping point on the fabric, followed by the robot’s successful execution of the grasp using the gripper.

\textbf{Inference Time (IT):} The duration required for the model to process the input data and generate a prediction.

These metrics are applied across four  real-world cloth manipulation tasks, including:

\textbf{1) Single Fold (SF):} One corner is pulled to its opposing corner.  

\textbf{2) Double Inward Fold (DIF):} Two opposing corners are folded to the center. 

\textbf{3) Double Triangle Fold (DTF):} Two sets of opposing corners are aligned with each other. 

\textbf{4) Four Corners Inward Fold (FCIF):} All corners are sequentially folded to the center.

\textbf{5) T-Shirt Fold (TSF):} The two sleeves of a t-shirt are folded to the center of the shirt, with the midpoint of the bottom folded to the center of the collar.

\textbf{6) Shorts Fold(ShF):} The legs of the shorts are first folded in half lengthwise, and then this fold is further halved to form a double-folded structure.

We used 18 samples from each of three fabric categories—towels, T-shirts, and shorts (see Fig. 7).

\begin{table}[t]
\centering
\renewcommand{\arraystretch}{1.5} % 调整行距
\caption{Evaluation results from crumpled initial configurations to smooth final states, selecting the highest score across all timesteps in the episode.}
\begin{tabular}{lccc}
\toprule
Method & Initial ISC(\%) & Finial ISC(\%) &$\Delta$ISC(\%)\\
\hline
Human & 63.1 & 98.8 & 35.7  \\
PickToPlace[11] & 64.6 & 95.6 & 31.0 \\
Ours &  \textbf{64.3}  &  \textbf{97.7}  &  \textbf{33.4}  \\
\bottomrule
\end{tabular}
\end{table}

\subsection{Comparison}
We compare our method to three other approaches: Lee et al. \cite{10}, PickToPlace \cite{11}, and FabricFlowNet \cite{15}. We selected these three baseline methods for their distinct approaches to cloth manipulation: 1) Lee et al. use offline RL for arbitrary goal cloth folding; 2) FabricFlowNet employs optical flow to predict grasping points from simulation-generated data for fabric folding; 3) The third method, PickToPlace, utilizes little human demonstration data for cloth manipulation and is considered the state-of-the-art (SOTA). The comparative analysis focuses on three distinct tasks: smoothing, folding, and the full task—which combines both smoothing and folding—with each task type undergoing 6 trials. All experiments were conducted on the same experimental platform developed in our study to ensure consistency across methods. Additionally, we modify FabricFlowNet to a single-arm variant. The training dataset comprised 800 images of various fabrics, including towels, T-shirts, and shorts. For model training, we set the learning rate to 0.001, used the Adam optimizer, and employed a batch size of 32. The experimental environment was controlled with consistent lighting and a fixed workspace size of 1m², ensuring fairness across all tests.

\begin{figure*}[t]
  \centering
  \includegraphics[width=1\linewidth]{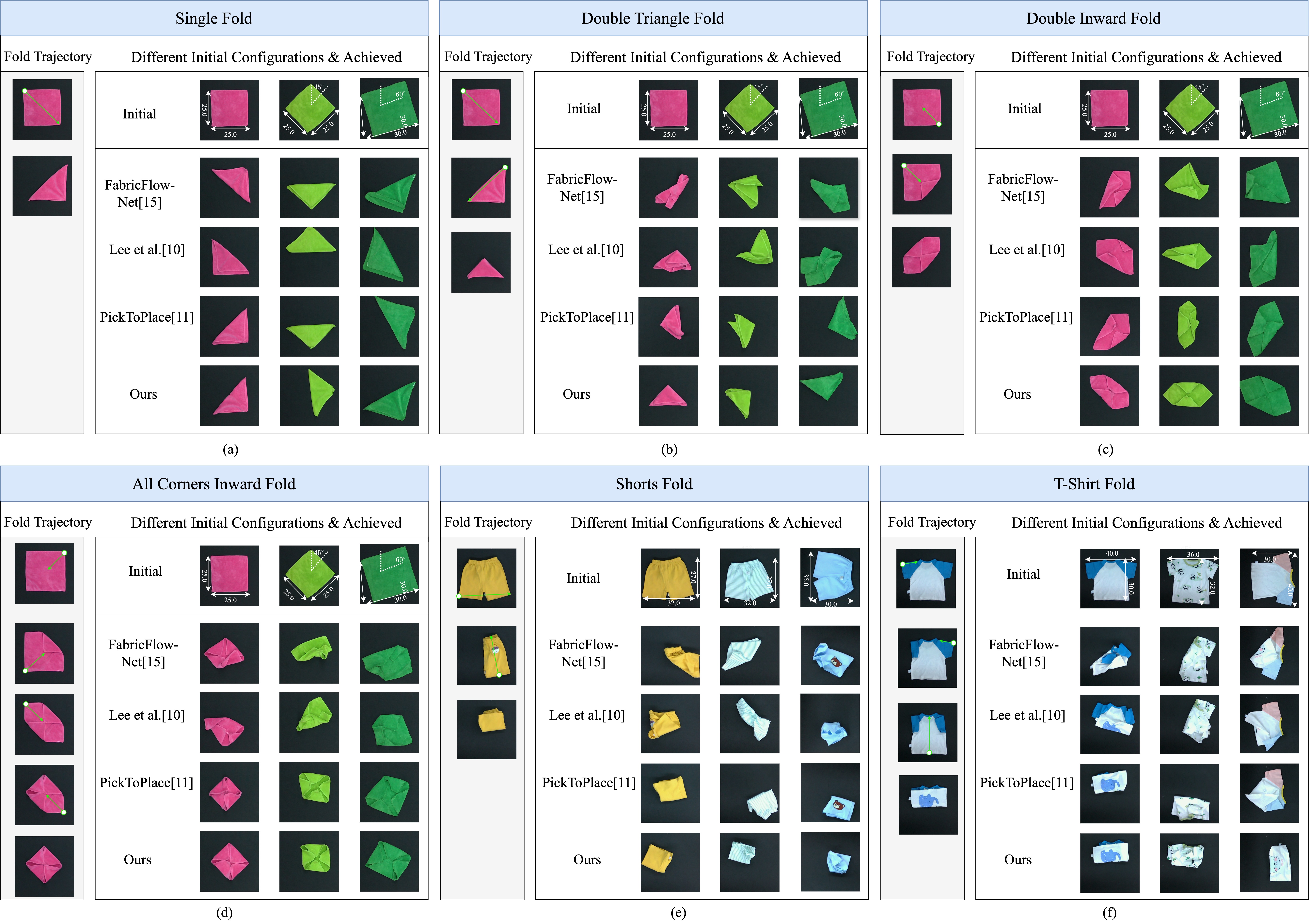}
  \caption{\textbf{Qualitative Comparison Results. } Our method successfully executes three sequential multi-step tasks for manipulating square cloths in a real-world setting: (a) Single Fold (b) Double Inward Fold, (c) Double Triangle Fold, (d) Four Corners Inward Fold,(e) Shorts Fold,(f) T-Shirt Fold.}
  \end{figure*}

\begin{table*}[t]
\centering
\setlength{\tabcolsep}{2.5pt} % Adjust column spacing
\renewcommand{\arraystretch}{1.5}
\caption{Results of Our Method and Three Baselines for Folding Across All Tasks. The Best Results are Bolded.}
\begin{tabular}{lccccccccccccccccccccccc}
\toprule
\multirow{2}{*}{Method}
& \multicolumn{3}{c}{SF} & \multicolumn{3}{c}{DIF} & \multicolumn{3}{c}{DTF} & \multicolumn{3}{c}{FCIF} & \multicolumn{3}{c}{ShF} & \multicolumn{3}{c}{TSF}
& \multirow{2}{*}{GA(\%)} & \multirow{2}{*}{IT(ms)} \\
\cmidrule(lr){2-4} \cmidrule(lr){5-7} \cmidrule(lr){8-10} \cmidrule(lr){11-13} \cmidrule(lr){14-16} \cmidrule(lr){17-19}
& A & S(\%) & IoU(\%) & A & S(\%) & IoU(\%) & A & S(\%) & IoU (\%)& A & S(\%) & IoU(\%) & A & S(\%) & IoU(\%) & A & S(\%) & IoU(\%) & \\
\hline
Lee et al.~\cite{10} & 1 & 83.3 & 86.4 & 2 & 50.0 & 81.0 & 2 & 50.0 & 80.9 & 4 & 50.0 & 80.7 & 2 & 66.7 & 80.5 & 3 & 0 & 76.6 & 57.1 & \textbf{0.331} \\
FabricFlowNet~\cite{15} & 1 & 66.7 & 84.5 & 2 & 66.7 & 83.9 & 2 & 33.3 & 78.8 & 4 & 33.3 & 77.1 & 2 & 50.0 & 79.2 & 3 & 16.7 & 77.7 & 64.2 & 1.356 \\
PickToPlace~\cite{11} & 1 & 100.0 & 88.3 & 2 & 83.3 & 86.3 & 2 & 83.3 & 85.7 & 4 & 66.7 & 83.5 & 2 & 66.7 & 82.9 & 3 & 50.0 & 83.6 & 85.7 & 6.613 \\
\textbf{Ours} & 1 & \textbf{100.0} & \textbf{89.1} & 2 & \textbf{100.0} & \textbf{87.6} & 2 & \textbf{83.3} & \textbf{86.1} & 4 & \textbf{83.3} & \textbf{85.7} & 2 & \textbf{83.3} & \textbf{85.3} & 3 & \textbf{66.7} & \textbf{84.8} & \textbf{92.8} & 13.86 \\
\bottomrule
\end{tabular}
\end{table*}

\textbf{For the smoothing,} We initiated the fabrics with varying degrees of crumpling and conducted 6 trials in $\mathit{SF}$ task to evaluate performance. The results, presented in Table I, demonstrate that our method achieved an impressive intermediate Smooth Coverage (ISC) of 97.7\%, closely approaching the near-perfect benchmark of 98.8\% set by human performance. This high level of accuracy underscores the effectiveness of our approach in smoothing fabrics across a wide range of initial conditions. Notably, our method produced a significant increase in fabric smoothness, with a $\Delta$$\mathit{ISC}$ of 33.4\%, outperforming the PickToPlace[11], which achieved a $\Delta$$\mathit{ISC}$ of 31.1\%. This superior performance can be attributed to our use of a two-stream architecture, which enhances the model’s ability to handle the complexity of fabric textures and crumpling patterns. (discussed in detail in the ablation section).

 \textbf{For cloth folding,}   Table II presents the quantitative results, while Fig. 8 shows some representative qualitative results. As observed, all methods begin with the clothes in a flattened state and consider the outcome successful if the final state visually aligns with the target image. As shown in Table II, our method significantly outperforms the baseline approaches across all tasks. Although the inference time (IT) of our method is slightly slower than that of the baseline approaches, it remains acceptable for real-time applications.

  \begin{figure*}[t]
    \centering
    \includegraphics[width=1\linewidth]{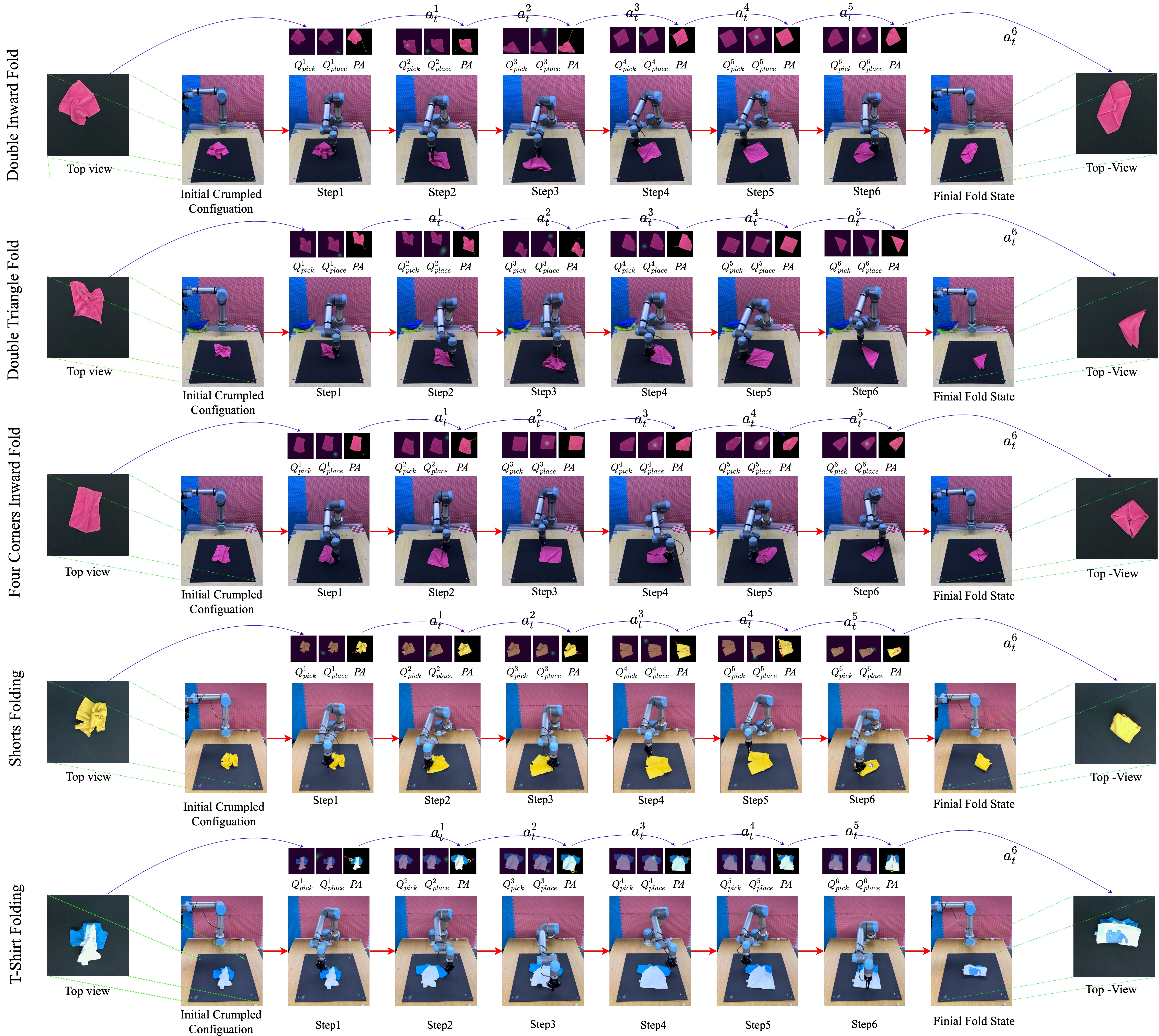}
    \caption{\textbf{Sequential Demonstration of Folding Arbitrary Crumpled Cloth Across Three Distinct Tasks.} Each task consists of two rows: the first row presents the top-view operation sequence captured by the overhead camera, while the second row displays the side-view operation sequence captured by the side camera. $Q_{\text {pick}}$ represents the predicted pick heatmap, $Q_{\text {place}}$ represents the predicted place heatmap, $PA$ represents the predicted action map, and $a_{t}$  represents the action pair.}
    \end{figure*}

We achieved perfect success rates of 100\% for the $SF$ and $\mathit{DIF}$ tasks, which involved fewer actions (A) and single-layer cloth manipulation. In contrast, the success rates for the \textit{DTF} and \textit{ShF} tasks decreased to 83.3\%, primarily due to the increased complexity of handling multi-layer cloth. The success rates for the \textit{FCIF} task also dropped to 83.3\%, which can be attributed to the higher number of actions (A) required. For the \textit{TSF} task, the success rate further decreased to 66.7\%, reflecting the additional challenge of handling multi-layer and irregularly shaped cloth. Videos of the robot executions are available on our project website.

\textbf{For the full task,} We classify task difficulty into three levels—easy, medium, and hard—based on fabric coverage. Specifically, tasks are deemed easy when fabric coverage exceeds 0.7, medium when it falls between 0.5 and 0.7, and hard when it is below 0.5. Table III and Fig. 9 present both quantitative and qualitative results for our method of folding arbitrarily crumpled cloth across several tasks.

In the $\mathit{SF}$ task, our method achieved a perfect success rate of 12/12, with an average success of 6/6. The IoU score was 88.2\%, close to human performance (93.3\%), and the ISC was 98.7\%, matching human-level results (99.0\%). In the $\mathit{DIF}$ task, we achieved a success rate of 11/12, with an average success of 5.5/6. The IoU score was 88.0\%, comparable to human performance (92.1\%), and the ISC was 98.4\%, slightly lower than human performance (99.2\%), but still close.
In the more challenging $\mathit{DTF}$ task, we achieved a success rate of 11/12, with an average success of 5.5/6. However, the IoU score decreased to 86.9\%, and the ISC dropped to 98.9\%. This decline can be attributed to the added complexity of handling double-layer fabric during the second fold, which poses additional challenges in manipulating multiple layers of cloth. In the $\mathit{FCIF}$ task, our method succeeded in 6/12 trials, with an average success of 3/6. The IoU score was 86.9\%, and the ISC was 98.9\%. The lower performance in this task is primarily due to the multi-step nature of the folding process. A small error during any step can trigger re-flattening, preventing the model from completing the final folded state and reducing the overall success rate.
   \begin{figure*}[t]
    \centering
    \includegraphics[width=0.95\linewidth]{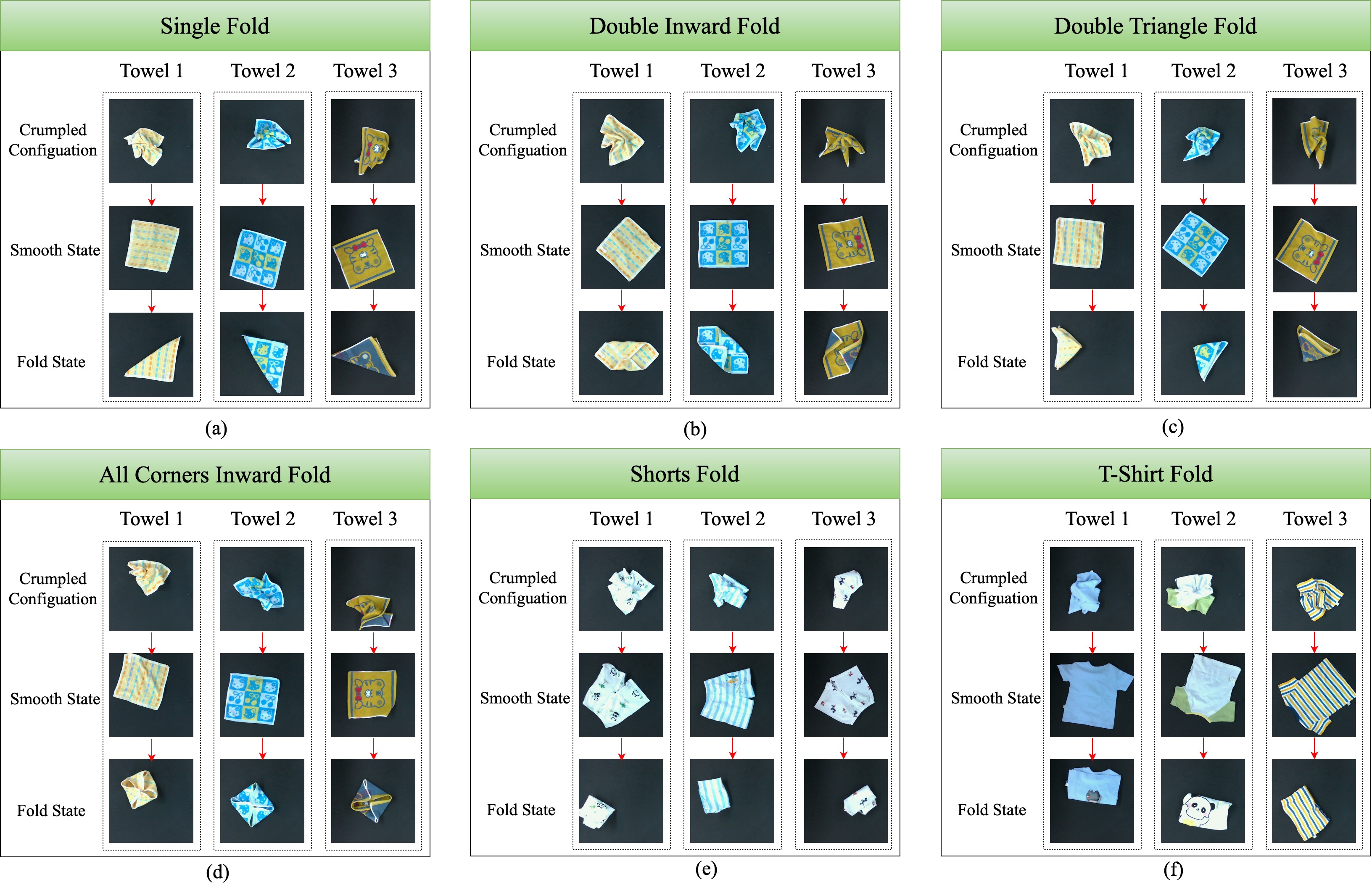}
    \caption{\textbf{Unseen Cloth Instances for Testing.}  Our method generalizes to a variety of cloth appearances and textures. We demonstrate the initial crumpled state, the smooth state, and the final configurations from successful folding episode.} 
    \end{figure*}

For the multi-layer fabric tasks, such as $\mathit{ShF}$ and $\mathit{STF}$, where the folding targets are irregular, our method achieved success rates of 9/12 and 8/12, with average successes of 4.5/6 and 4/6, respectively. The IoU scores were 86.9\% for $\mathit{ShF}$ and 84.7\% for $\mathit{STF}$, and the ISC values were 98.9\% for $\mathit{ShF}$ and 97.3\% for $\mathit{STF}$. The irregular nature of the folding targets, combined with the complexity of handling multi-layer fabrics, led to a slight reduction in fold accuracy and success rates compared to simpler tasks.

In summary, our method consistently outperformed the PickToPlace baseline across all tasks, achieving near-human-level performance in the $\mathit{SF}$ and $\mathit{DIF}$ tasks. However, performance was less robust in more challenging tasks, such as $\mathit{TSF}$ and $\mathit{FCIF}$, where the IoU and ISC scores were lower. Future work will focus on improving the robustness of our method in these more challenging scenarios, particularly through more effective handling of multi-step folding and irregular targets.

\textbf{Generalization to Unseen cloth,} our method, trained on both depth and RGB images, exhibits strong generalization across various visual properties of cloth, ensuring robustness to both color and texture variations. This two-stream approach enhances the model's ability to adapt to a wide range of material properties such as thickness, texture, and shape. We conducted extensive testing on diverse cloth types, with the results detailed in Fig. 10. These findings demonstrate that our model not only generalizes well to different visual appearances but is also effective across variations in material properties, underscoring its potential for broad practical application in cloth manipulation tasks.

\subsection{Ablation}

To assess the effectiveness of our method's different components, we designed two ablation studies to evaluate the contributions of each of the three novel aspects.

\textbf{W/O PIC (Without Pick-Conditions):} the approach utilizes Pick Conditions to enhance the accuracy of place predictions.

\textbf{W/O SPM (Without Spatial Module):} without learning a visible connectivity graph.

Each ablation study involved 6 trials, with quantitative results presented in Table IV. The findings clearly indicate that each component plays a significant role in enhancing the method’s performance. Specifically, the Pick-Conditions are critical for improving placement accuracy by guiding the agent in predicting optimal pick locations. Meanwhile, the spatial module, through the visible connectivity graph, enables the agent to better understand the spatial relationships of the cloth, particularly in the presence of self-occlusion. The combination of these two components leads to a substantial increase in accuracy compared to the baseline. For the $\mathit{FCIF}$ and $\mathit{TSF}$ tasks, the success rate decreased to 66.7\%, reflecting the increased number of required actions and the additional challenges posed by multi-layer configurations and irregularly shaped cloth.

\begin{table*}[t]
\centering
\setlength{\tabcolsep}{2.5pt}
\renewcommand{\arraystretch}{1.5} % 调整行距
\caption{Comparison of Methods Across Different Cloth Difficulty Levels: Simple (Easy \& Medium) and Hard.}
\begin{tabular}{lccccccccccccccccccccccc}
\toprule
Task  & Method & \multicolumn{3}{c}{SF} & \multicolumn{3}{c}{DIF} & \multicolumn{3}{c}{DTF} & \multicolumn{3}{c}{FCIF} & \multicolumn{3}{c}{ShF} & \multicolumn{3}{c}{TSF} \\
\cmidrule(lr){3-5} \cmidrule(lr){6-8} \cmidrule(lr){9-11} \cmidrule(lr){12-14} \cmidrule(lr){15-17} \cmidrule(lr){18-20}
& & S & IoU(\%) & ISC(\%) & S & IoU(\%) & ISC(\%) & S & IoU(\%) & ISC(\%) & S & IoU(\%) & ISC(\%) & S & IoU(\%) & ISC(\%) & S & IoU(\%) & ISC(\%) \\
\hline
    \multirow{3}{*}{Simple} 
     & Human & 6 & 93.3 & 99.0 & 6 & 92.1 & 99.2 & 6 & 92.7 & 99.4 & 6 & 91.8 & 98.6 & 6 & 89.6 & 98.1 & 6 & 85.0 & 98.3 \\
     & PickToPlace[11] & 6 & 88.1 & 97.4 & 5 & 86.3 & 97.9 & 6 & 86.1 & 97.1 & 2 & 77.8 & 96.7 & 4 & 84.4 & 96.3 & 2 & 76.6 & 95.2 \\
     & \textbf{Ours} & \textbf{6} & \textbf{88.2} & \textbf{98.7} & \textbf{6} & \textbf{88.0} & \textbf{98.4} & \textbf{6} & \textbf{86.9} & \textbf{98.9} & \textbf{3} & \textbf{79.9} & \textbf{98.1} & \textbf{5} & \textbf{85.2} & \textbf{97.1} & \textbf{4} & \textbf{81.5} & \textbf{97.8} \\
     \hline
    \multirow{3}{*}{Hard}
     & Human & 6 & 93.7 & 98.8 & 6 & 91.9 & 99.5 & 6 & 93.0 & 98.7 & 6 & 91.0 & 99.2 & 6 & 89.8 & 98.3 & 6 & 84.5 & 98.9 \\
     & PickToPlace[11] & 6 & 86.9 & 95.6 & 5 & 85.4 & 95.9 & 4 & 82.7 & 95.4 & 2 & 76.1 & 94.6& 3 & 80.7 & 94.5 &  2& 74.7 & 93.9 \\
     & \textbf{Ours} & \textbf{6} & \textbf{88.0} & \textbf{97.7} & \textbf{5} & \textbf{85.7} & \textbf{98.0} & \textbf{5} & \textbf{84.7} & \textbf{97.3} & \textbf{3} & \textbf{78.7} & \textbf{97.2} & \textbf{4} & \textbf{82.4} & \textbf{95.9} & \textbf{4} & \textbf{80.2} & \textbf{95.4} \\
     \hline
     \multirow{3}{*}{Mean}
     & Human & 6 & 93.5 & 98.9 & 6 & 92.0 & 99.4 & 6 & 92.9 & 99.1 & 6 & 91.4 & 98.9 & 6 & 89.7 & 98.2 & 6 & 84.8 & 98.6 \\
     & PickToPlace[11] & 6 &87.5 & 96.5 & 5 & 85.9 & 96.9 & 5 & 84.4 & 96.3 & 2 & 76.9 & 95.7 & 4 & 82.3 & 95.4 & 2 & 75.7 & 94.6 \\
     & \textbf{Ours} & \textbf{6} & \textbf{88.1} & \textbf{98.2} & \textbf{5.5} & \textbf{86.7} & \textbf{98.2} & \textbf{5.5} & \textbf{85.8} & \textbf{98.1} & \textbf{3} & \textbf{78.4} & \textbf{97.7} & \textbf{4.5} & \textbf{83.8} & \textbf{96.5} & \textbf{4} & \textbf{80.9} & \textbf{96.6} \\
    \bottomrule
    \end{tabular}
  \end{table*}

\begin{table}[t]
  \centering
  \setlength{\tabcolsep}{4pt} 
  \renewcommand{\arraystretch}{1.5} % 调整行距
  \caption{results of ablations.}
  \begin{tabular}{lcccccc}
  \toprule
  Condition & SF(\%) & DIF(\%) & DTF(\%) & FCIF(\%) & ShF(\%) & TS(\%) \\
  \hline
  W/O PIC & 66.7 & 50.0 & 50.0 & 16.7 & 33.3 & 33.3 \\
  W/O SPM & 83.3 & 66.7 & 50.0 & 50.0 & 66.7 & 33.3 \\
 \textbf{Ours} & \textbf{100.0} & \textbf{100.0} & \textbf{83.3} & \textbf{66.7} & \textbf{83.3} & \textbf{66.7}  \\
  \bottomrule
  \end{tabular}
  
  \end{table}

\section{Discussion}

\textbf{Summary.} In this work, we present SSFold, a novel framework for cloth folding tasks using a two-stream architecture with sequential and spatial pathways. It effectively handles various fabric types and transforms crumpled cloth into the desired folded states, achieving high success rates across tasks. Validated with a UR5 robotic arm in real-world environments, SSFold shows strong generalization and robustness across different fabric types and states.

\textbf{The Scalability. }The SSFold framework is well-suited for scalable deployment in industrial environments such as textile manufacturing. By leveraging a YOLOv10-based system for hand tracking and key-point detection, our approach efficiently collects high-fidelity demonstration data from diverse, real-world settings, including textile factories. To address noise and inaccuracies, we implemented a data quality control framework with trajectory correction and real-time visualization (see Fig.11). A nearest-neighbor optimization algorithm adjusts inaccurate pick-and-place points to align with the cloth region, with continuous real-time feedback ensuring high-quality data for training. Furthermore, the framework’s adaptability extends to a variety of fabric types with varying elasticity, stiffness, and texture. The system is also designed for dual-arm configurations, improving efficiency for large-scale tasks such as folding larger fabrics. SSFold's platform-agnostic design allows easy integration with different robots, while its use of affordable hand-tracking reduces costs. This makes SSFold a cost-effective option for industrial cloth manipulation, easing adoption for factories.

\begin{figure}[t]
  \centering
\centering
\includegraphics[width=1\linewidth]{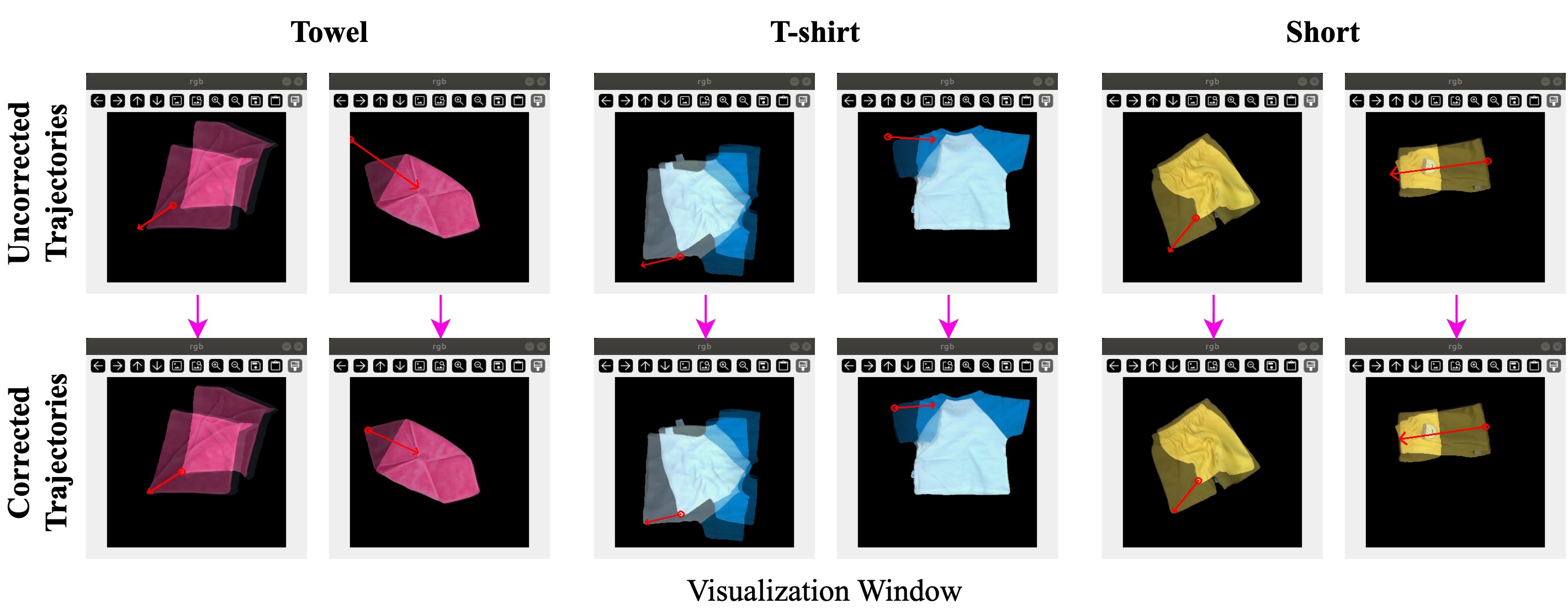}
%\caption{fig2}
\caption{\textbf{Data Preprocessing.} Preprocessing of human demonstration data to correct noise and inaccuracies.}
\label{}
\end{figure}

\textbf{Limitations and Future Work. }Despite the promising results, there are notable challenges when the system is applied to more complex tasks, such as the double triangle fold, T-shirt fold, and the four corners inward fold. These challenges stem primarily from issues related to multi-layer cloth manipulation and the accumulation of errors in multi-step tasks. The occlusion caused by overlapping layers impedes accurate state estimation, and the current top-down grasping approach limits the system’s ability to effectively distinguish between single-layer and multi-layer grasp points. Furthermore, in multi-step tasks, errors from earlier steps can propagate, negatively affecting the overall success rate.  Future work will focus on addressing these limitations by integrating multi-view cameras to reduce occlusions, incorporating tactile feedback to improve layer differentiation, and optimizing gripper design and orientation for better handling of complex cloth configurations. We also plan to explore advanced cloth dynamics modeling and improve system adaptability for diverse environments.

\section{Conclusion}
We introduced SSFold, a novel YOLOv10-based method for robotic cloth manipulation that efficiently utilizes human demonstration data. Our two-stream architecture successfully combines smoothing and folding tasks, adapting to various cloth types and states. Validated on a UR5 robotic arm, our method outperforms existing techniques, demonstrating strong generalization across different cloth properties. This approach not only addresses challenges like the sim-to-real gap and complex cloth dynamics but also enhances the feasibility of deploying robotic cloth manipulation in practical applications. Future research will focus on improving sim-to-real transfer and further optimizing the handling of complex cloth dynamics, aiming to broaden the practical deployment of robotic systems in cloth manipulation.

\bibliographystyle{IEEEtran}  % 选择IEEE样式
\bibliography{references.bib}     % 指向 .bib 文件

% Generated by IEEEtran.bst, version: 1.14 (2015/08/26)
\begin{thebibliography}{10}
\providecommand{\url}[1]{#1}
\csname url@samestyle\endcsname
\providecommand{\newblock}{\relax}
\providecommand{\bibinfo}[2]{#2}
\providecommand{\BIBentrySTDinterwordspacing}{\spaceskip=0pt\relax}
\providecommand{\BIBentryALTinterwordstretchfactor}{4}
\providecommand{\BIBentryALTinterwordspacing}{\spaceskip=\fontdimen2\font plus
\BIBentryALTinterwordstretchfactor\fontdimen3\font minus \fontdimen4\font\relax}
\providecommand{\BIBforeignlanguage}[2]{{%
\expandafter\ifx\csname l@#1\endcsname\relax
\typeout{** WARNING: IEEEtran.bst: No hyphenation pattern has been}%
\typeout{** loaded for the language `#1'. Using the pattern for}%
\typeout{** the default language instead.}%
\else
\language=\csname l@#1\endcsname
\fi
#2}}
\providecommand{\BIBdecl}{\relax}
\BIBdecl

\bibitem{1}
N.-Q. Gu, R.~He, and L.~Yu, ``Learning to unfold garment effectively into oriented direction,'' \emph{IEEE Robotics and Automation Letters}, 2023.

\bibitem{2}
H.~Shehawy, D.~Pareyson, V.~Caruso, S.~{De Bernardi}, A.~M. Zanchettin, and P.~Rocco, ``Flattening and folding towels with a single-arm robot based on reinforcement learning,'' \emph{Robotics and Autonomous Systems}, vol. 169, p. 104506, 2023.

\bibitem{3}
B.~Thananjeyan, A.~Balakrishna, U.~Rosolia, F.~Li, R.~McAllister, J.~E. Gonzalez, S.~Levine, F.~Borrelli, and K.~Goldberg, ``Safety augmented value estimation from demonstrations (saved): Safe deep model-based rl for sparse cost robotic tasks,'' \emph{IEEE Robotics and Automation Letters}, vol.~5, no.~2, pp. 3612--3619, 2020.

\bibitem{4}
E.~Torgerson and F.~W. Paul, ``Vision-guided robotic fabric manipulation for apparel manufacturing,'' \emph{IEEE Control Systems Magazine}, vol.~8, no.~1, pp. 14--20, 1988.

\bibitem{5}
N.~Essahbi, B.~C. Bouzgarrou, and G.~Gogu, ``Soft material modeling for robotic manipulation,'' \emph{Applied Mechanics and Materials}, vol. 162, pp. 184--193, 2012.

\bibitem{6}
V.~Raval, E.~Zhao, H.~Zhang, S.~Nikolaidis, and D.~Seita, ``Gpt-fabric: Folding and smoothing fabric by leveraging pre-trained foundation models,'' \emph{arXiv preprint arXiv:2406.09640}, 2024.

\bibitem{7}
X.~Lin, Y.~Wang, Z.~Huang, and D.~Held, ``Learning visible connectivity dynamics for cloth smoothing,'' in \emph{Conference on Robot Learning}.\hskip 1em plus 0.5em minus 0.4em\relax PMLR, 2022, pp. 256--266.

\bibitem{8}
A.~Canberk, C.~Chi, H.~Ha, B.~Burchfiel, E.~Cousineau, S.~Feng, and S.~Song, ``Cloth funnels: Canonicalized-alignment for multi-purpose garment manipulation,'' in \emph{2023 IEEE International Conference on Robotics and Automation (ICRA)}.\hskip 1em plus 0.5em minus 0.4em\relax IEEE, 2023, pp. 5872--5879.

\bibitem{9}
Y.~Li, J.~Wu, R.~Tedrake, J.~B. Tenenbaum, and A.~Torralba, ``Learning particle dynamics for manipulating rigid bodies, deformable objects, and fluids,'' \emph{arXiv preprint arXiv:1810.01566}, 2018.

\bibitem{10}
R.~Lee, D.~Ward, V.~Dasagi, A.~Cosgun, J.~Leitner, and P.~Corke, ``Learning arbitrary-goal fabric folding with one hour of real robot experience,'' in \emph{Conference on Robot Learning}.\hskip 1em plus 0.5em minus 0.4em\relax PMLR, 2021, pp. 2317--2327.

\bibitem{11}
R.~Lee, J.~Abou-Chakra, F.~Zhang, and P.~Corke, ``Learning fabric manipulation in the real world with human videos,'' in \emph{2024 IEEE International Conference on Robotics and Automation (ICRA)}.\hskip 1em plus 0.5em minus 0.4em\relax IEEE, 2024, pp. 3124--3130.

\bibitem{12}
A.~Choi, D.~Tong, D.~Terzopoulos, J.~Joo, and M.~K. Jawed, ``Learning neural force manifolds for sim2real robotic symmetrical paper folding,'' \emph{IEEE Transactions on Automation Science and Engineering}, 2024.

\bibitem{13}
A.~Wang, H.~Chen, L.~Liu, K.~Chen, Z.~Lin, J.~Han, and G.~Ding, ``Yolov10: Real-time end-to-end object detection,'' \emph{arXiv preprint arXiv:2405.14458}, 2024.

\bibitem{14}
Y.~Wu, W.~Yan, T.~Kurutach, L.~Pinto, and P.~Abbeel, ``Learning to manipulate deformable objects without demonstrations,'' \emph{arXiv preprint arXiv:1910.13439}, 2019.

\bibitem{15}
T.~Weng, S.~M. Bajracharya, Y.~Wang, K.~Agrawal, and D.~Held, ``Fabricflownet: Bimanual cloth manipulation with a flow-based policy,'' in \emph{Conference on Robot Learning}.\hskip 1em plus 0.5em minus 0.4em\relax PMLR, 2022, pp. 192--202.

\bibitem{16}
G.~A. Odesanmi, Q.~Wang, and J.~Mai, ``Skill learning framework for human--robot interaction and manipulation tasks,'' \emph{Robotics and Computer-Integrated Manufacturing}, vol.~79, p. 102444, 2023.

\bibitem{17}
Y.~Deng, K.~Mo, C.~Xia, and X.~Wang, ``Learning language-conditioned deformable object manipulation with graph dynamics,'' in \emph{2024 IEEE International Conference on Robotics and Automation (ICRA)}.\hskip 1em plus 0.5em minus 0.4em\relax IEEE, 2024, pp. 7508--7514.

\bibitem{18}
Q.~Gao, Z.~Ju, Y.~Chen, Q.~Wang, Y.~Zhao, and S.~Lai, ``Parallel dual-hand detection by using hand and body features for robot teleoperation,'' \emph{IEEE Transactions on Human-Machine Systems}, vol.~53, no.~2, pp. 417--426, 2023.

\bibitem{19}
A.~C. Bavelos, E.~Anastasiou, N.~Dimitropoulos, G.~Michalos, and S.~Makris, ``Virtual reality-based dynamic scene recreation and robot teleoperation for hazardous environments,'' \emph{Computer-Aided Civil and Infrastructure Engineering}, 2024.

\bibitem{20}
S.~Elliott, Z.~Xu, and M.~Cakmak, ``Learning generalizable surface cleaning actions from demonstration,'' in \emph{2017 26th IEEE international symposium on robot and human interactive communication (RO-MAN)}.\hskip 1em plus 0.5em minus 0.4em\relax IEEE, 2017, pp. 993--999.

\bibitem{21}
K.-I. Yoon, D.-K. Ko, and S.-C. Lim, ``Real-time video prediction using gans with guidance information for time-delayed robot teleoperation,'' \emph{International Journal of Control, Automation and Systems}, vol.~21, no.~7, pp. 2387--2397, 2023.

\bibitem{22}
X.~Sun, J.~Li, A.~V. Kovalenko, W.~Feng, and Y.~Ou, ``Integrating reinforcement learning and learning from demonstrations to learn nonprehensile manipulation,'' \emph{IEEE Transactions on Automation Science and Engineering}, vol.~20, no.~3, pp. 1735--1744, 2023.

\bibitem{23}
W.~Wang, C.~Zeng, H.~Zhan, and C.~Yang, ``A novel robust imitation learning framework for complex skills with limited demonstrations,'' \emph{IEEE Transactions on Automation Science and Engineering}, pp. 1--13, 2024.

\bibitem{24}
S.~Kumar, J.~Zamora, N.~Hansen, R.~Jangir, and X.~Wang, ``Graph inverse reinforcement learning from diverse videos,'' in \emph{Conference on Robot Learning}.\hskip 1em plus 0.5em minus 0.4em\relax PMLR, 2023, pp. 55--66.

\bibitem{25}
H.~Ha and S.~Song, ``Flingbot: The unreasonable effectiveness of dynamic manipulation for cloth unfolding,'' in \emph{Conference on Robot Learning}.\hskip 1em plus 0.5em minus 0.4em\relax PMLR, 2022, pp. 24--33.

\bibitem{26}
D.~Seita, N.~Jamali, M.~Laskey, A.~K. Tanwani, R.~Berenstein, P.~Baskaran, S.~Iba, J.~Canny, and K.~Goldberg, ``Deep transfer learning of pick points on fabric for robot bed-making,'' in \emph{The International Symposium of Robotics Research}.\hskip 1em plus 0.5em minus 0.4em\relax Springer, 2019, pp. 275--290.

\bibitem{27}
L.~Sun, G.~Aragon-Camarasa, S.~Rogers, and J.~P. Siebert, ``Accurate garment surface analysis using an active stereo robot head with application to dual-arm flattening,'' in \emph{2015 IEEE international conference on robotics and automation (ICRA)}.\hskip 1em plus 0.5em minus 0.4em\relax IEEE, 2015, pp. 185--192.

\bibitem{28}
J.~Qian, T.~Weng, L.~Zhang, B.~Okorn, and D.~Held, ``Cloth region segmentation for robust grasp selection,'' in \emph{2020 IEEE/RSJ International Conference on Intelligent Robots and Systems (IROS)}.\hskip 1em plus 0.5em minus 0.4em\relax IEEE, 2020, pp. 9553--9560.

\bibitem{29}
L.~Yang, Y.~Li, and L.~Chen, ``Clothppo: A proximal policy optimization enhancing framework for robotic cloth manipulation with observation-aligned action spaces,'' \emph{arXiv preprint arXiv:2405.04549}, 2024.

\bibitem{30}
R.~Wu, C.~Ning, and H.~Dong, ``Learning foresightful dense visual affordance for deformable object manipulation,'' in \emph{2023 IEEE/CVF International Conference on Computer Vision (ICCV)}, 2023, pp. 10\,913--10\,922.

\bibitem{31}
D.~Blanco-Mulero, G.~Alcan, F.~J. Abu-Dakka, and V.~Kyrki, ``Qdp: Learning to sequentially optimise quasi-static and dynamic manipulation primitives for robotic cloth manipulation,'' in \emph{2023 IEEE/RSJ International Conference on Intelligent Robots and Systems (IROS)}.\hskip 1em plus 0.5em minus 0.4em\relax IEEE, 2023, pp. 984--991.

\bibitem{32}
X.~Lin, Y.~Wang, J.~Olkin, and D.~Held, ``Softgym: Benchmarking deep reinforcement learning for deformable object manipulation,'' in \emph{Conference on Robot Learning}.\hskip 1em plus 0.5em minus 0.4em\relax PMLR, 2021, pp. 432--448.

\bibitem{33}
Z.~Huang, X.~Lin, and D.~Held, ``Mesh-based dynamics with occlusion reasoning for cloth manipulation,'' \emph{arXiv preprint arXiv:2206.02881}, 2022.

\bibitem{34}
S.~Wang, R.~Papallas, M.~Leonetti, and M.~Dogar, ``Goal-conditioned action space reduction for deformable object manipulation,'' in \emph{2023 IEEE International Conference on Robotics and Automation (ICRA)}, 2023, pp. 3623--3630.

\bibitem{35}
K.~Mo, C.~Xia, X.~Wang, Y.~Deng, X.~Gao, and B.~Liang, ``Foldsformer: Learning sequential multi-step cloth manipulation with space-time attention,'' \emph{IEEE Robotics and Automation Letters}, vol.~8, no.~2, pp. 760--767, 2023.

\bibitem{36}
R.~Wu, H.~Lu, Y.~Wang, Y.~Wang, and H.~Dong, ``Unigarmentmanip: A unified framework for category-level garment manipulation via dense visual correspondence,'' in \emph{Proceedings of the IEEE/CVF Conference on Computer Vision and Pattern Recognition (CVPR)}, June 2024.

\bibitem{37}
P.~Zhou, J.~Qi, A.~Duan, S.~Huo, Z.~Wu, and D.~Navarro-Alarcon, ``Imitating tool-based garment folding from a single visual observation using hand-object graph dynamics,'' \emph{IEEE Transactions on Industrial Informatics}, vol.~20, no.~4, pp. 6245--6256, 2024.

\end{thebibliography}

\vspace{-1cm}
\begin{IEEEbiography}[{\includegraphics[width=1in,height=1.25in,clip,keepaspectratio]{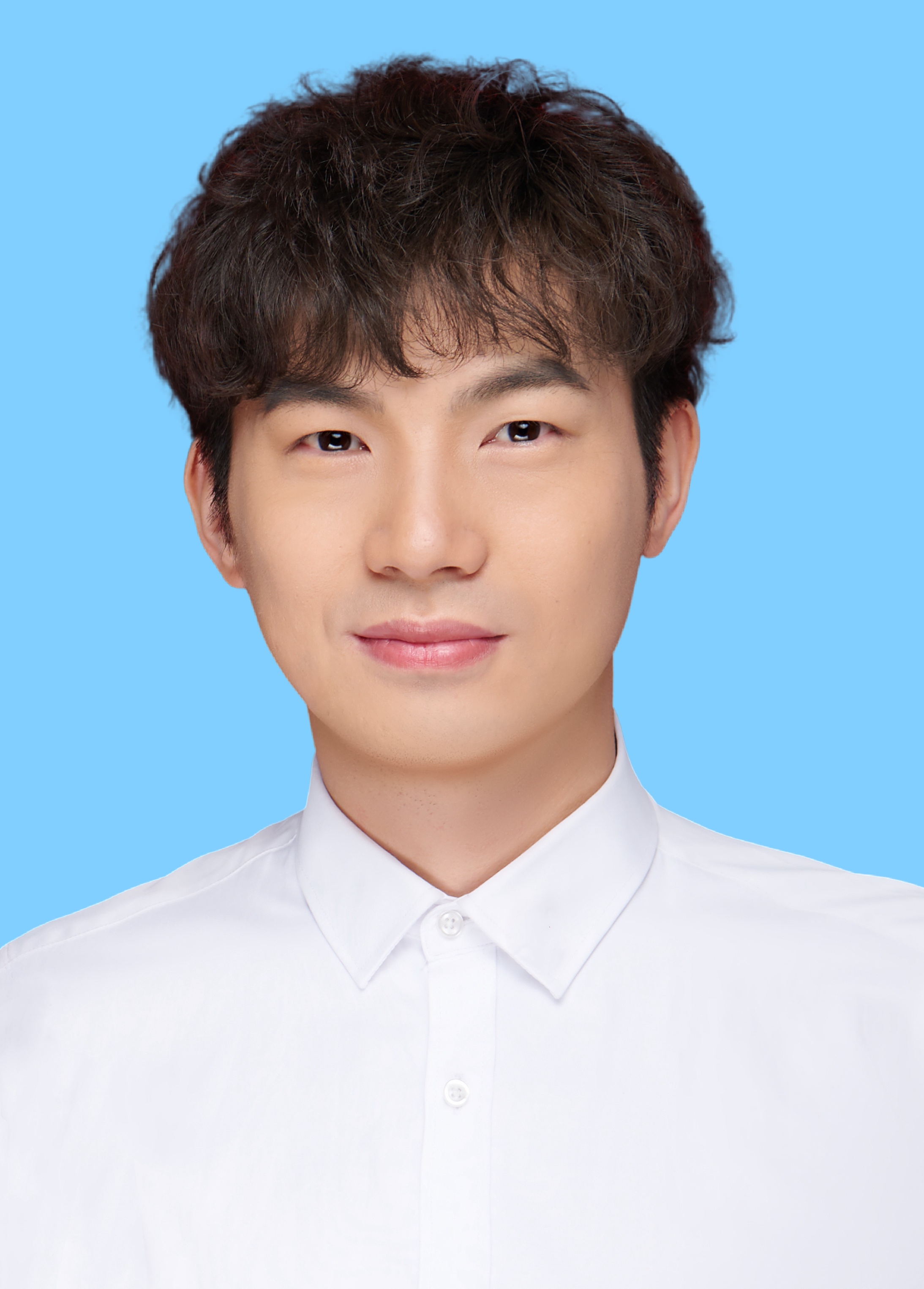}}]{Changshi Zhou}
  is currently pursuing the Ph.D. degree with the Shanghai Research Institute for Intelligent Autonomous Systems, Tongji University, China. His research interests include the manipulation and perception of flexible objects.
  \end{IEEEbiography}
  
\vspace{-1cm}
\begin{IEEEbiography}[{\includegraphics[width=1in,height=1.25in,clip,keepaspectratio]{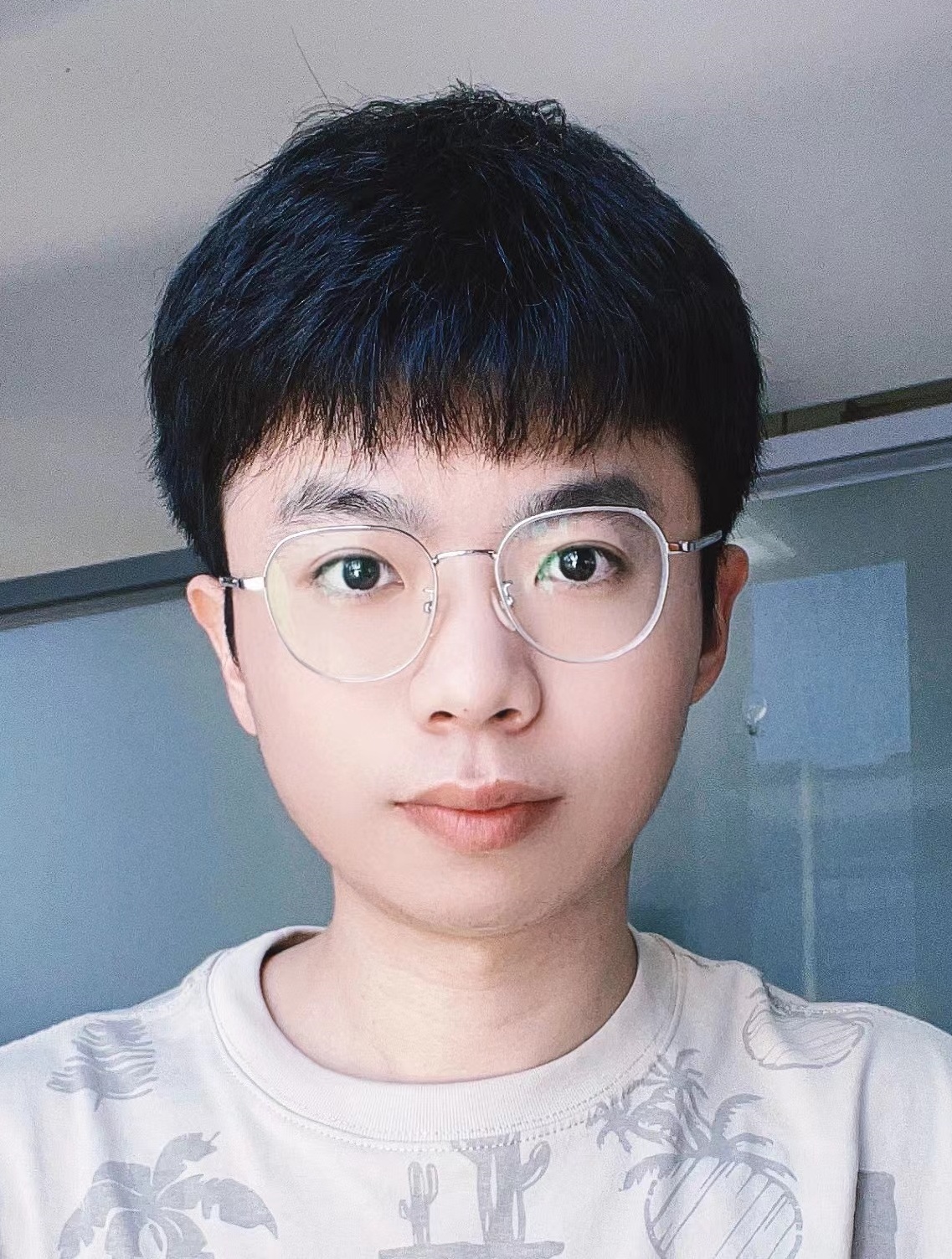}}]{Haichuan Xu} received his B.S. degree in information and computing science and an M.S. degree in applied mathematics from Hunan University of Technology, Zhuzhou, China, in 2019 and 2022, respectively. He is currently pursuing a Ph.D. degree in intelligent science and technology with Shanghai Research Institute for Intelligent Autonomous Systems, Tongji University, Shanghai, China. His current research interests include coordination control of multi-agent systems and its applications.
\end{IEEEbiography}
\vspace{-1cm}

\begin{IEEEbiography}[{\includegraphics[width=1in,height=1.25in,clip,keepaspectratio]{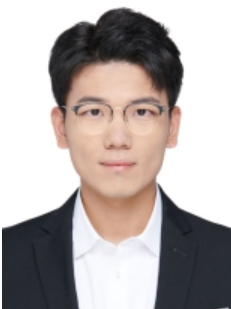}}]{Jiarui Hu}received the B.S. degree in mechanical engineering, Shandong University, Shandong, China, in 2021.He is currently working toward the Ph.D. degree in Control Science and Engineering from Tongji University, Shanghai, China. His research interests include tactile sensing and multimodal learning.
\end{IEEEbiography}

\vspace{-1cm}
\begin{IEEEbiography}[{\includegraphics[width=1in,height=1.25in,clip,keepaspectratio]{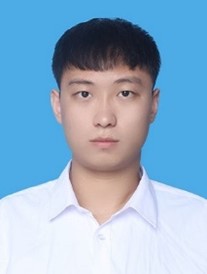}}]{Feng Luan}  is currently pursuing the Ph.D. degree with the Shanghai Research Institute for Intelligent Autonomous Systems, Tongji University, China. His research interests include tactile sensors, 3D reconstruction, and the design and control of intelligent robots.
\end{IEEEbiography}

\vspace{-1cm}
\begin{IEEEbiography}[{\includegraphics[width=1in,height=1.25in,clip,keepaspectratio]{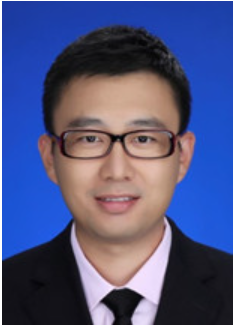}}]{Zhipeng Wang} received the M.S. degree from Zhejiang University, Hangzhou, China, in 2011, and the Ph.D. degree from Tongji University, Shanghai, China, in 2015. Between 2015 and 2018, he held postdoctoral research appointments with the College of Mechanical Engineering, Tongji University. He is currently a vice professor with the College of Electronics and Information Engineering, Tongji University. His current research interests include biped robot, mechatronics and dynamics.
\end{IEEEbiography}

\vspace{0cm}
\begin{IEEEbiography}[{\includegraphics[width=1in,height=1.25in,clip,keepaspectratio]{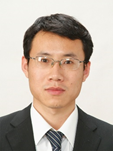}}]{Yanchao Dong } received the B.S. and M.S. degrees from Tongji University, Shanghai, China, in 2005 and 2008, respectively, and the Ph.D. degree in computer science from Kumamoto University, Kumamoto, Japan, in 2012. He was selected as one of the Chinese government’s overseas scholars to be a Ph.D. candidate at Kumamoto University in 2008. He is currently an Associate Professor with Tongji University and the Shanghai Research Institute for Intelligent Autonomous Systems, Shanghai. His research interests include visual SLAM, machine learning, and facial animation tracking.
\end{IEEEbiography}

\vspace{-12cm}
\begin{IEEEbiography}[{\includegraphics[width=1in,height=1.25in,clip,keepaspectratio]{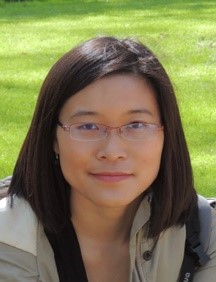}}]{Yanmin Zhou} received the M.S degree in control theory and control engineering from Tongji University, Shanghai, China, in 2011, and the Ph.D. degree in biomechanics from Cambridge University in 2015. She is currently a vice professor with the College of Electronics and Information Engineering, Tongji University. Her current research interests include bionics, the design and control of intelligent robot.
\end{IEEEbiography}

\vspace{-12cm}
\begin{IEEEbiography}[{\includegraphics[width=1in,height=1.25in,clip,keepaspectratio]{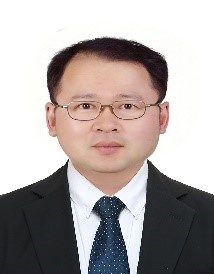}}]{Bin He} received the Ph.D. degree in mechanical and electronic control engineering from Zhejiang University, Hangzhou, China, in 2001, where he held postdoctoral research appointments with The State Key Lab of Fluid Power Transmission and Control, from 2001 and 2003. He is currently a Professor with the College of Electronics and Information Engineering, Tongji University, Shanghai, China. His current research interests include intelligent robot control, biomimetic microrobots, and wireless networks.
\end{IEEEbiography}

\newpage

\vfill

\end{document}